\documentclass[11pt,a4paper]{article}

\usepackage[utf8]{inputenc}
\usepackage[T1]{fontenc}
\usepackage[margin=2.5cm]{geometry}
\usepackage{amsmath,amssymb}
\usepackage{booktabs}
\usepackage{graphicx}
\usepackage{placeins}
\usepackage{tikz}
\usetikzlibrary{arrows.meta,positioning}
\usepackage{natbib}
\usepackage[hidelinks]{hyperref}
\hypersetup{
  pdftitle={VAIOM: Continuous-Input, Discrete-Output Decoder-Only Financial Sequence Modeling},
  pdfauthor={Yiming Ma and Xinyu Chen},
  pdfsubject={Probabilistic financial sequence modeling},
  pdfkeywords={financial time series, decoder-only Transformer, probabilistic return modeling, continuous input, discrete output}
}

\title{VAIOM: Continuous-Input, Discrete-Output Decoder-Only Financial Sequence Modeling}
\author{Yiming MA\thanks{These authors contributed equally.}\and Xinyu CHEN\footnotemark[1]}
\date{}

\begin{document}
\maketitle

\section*{Abstract}

Financial observations are continuous, heterogeneous and noisy, whereas decoder-only next-token models are usually built around discrete symbolic inputs. We introduce Vector-Input Autoregressive Inference for Ordinal-Return Modeling (VAIOM), a decoder-only Transformer for probabilistic next-return modeling on one-hour (1H) foreign-exchange (FX) bars. VAIOM separates input representation from output likelihood: continuous multivariate financial-event vectors preserve numerical structure at the input, while a categorical distribution over the next volatility-normalized return bucket supports cross-entropy training and likelihood evaluation.

The selected 0.9M Hybrid Continuous Input (HybridContIn) model combines continuous event features with categorical asset metadata and a Mixture-of-Market-States (MoMS) return head. It uses Gap, volatility-regime (VolReg) and Ordinal auxiliary objectives with full-sequence supervision (FullSeq). Models and preprocessing are fitted before 2024, selected on 2024H2 Validation and evaluated without refitting on two 2025 Test periods. The selected model outperforms train-fitted Frequency, Markov and identity-augmented single-bar Light Gradient Boosting Machine (LightGBM) baselines in both periods. Across three independent training seeds, every model beats the fixed LightGBM baseline in both Test halves. For the canonical checkpoint, paired gains over LightGBM are approximately 0.029--0.043 bits per event.

Validation experiments provide three main design findings. The evaluated continuous-input branch improves over the discrete-token branch while retaining the same categorical return objective; full-sequence autoregressive supervision improves over last-position training; and auxiliary representation shaping together with a mixture-structured return head improves return likelihood in controlled component comparisons. A supporting capacity study finds that the smallest evaluated complete architecture rung achieves the strongest Validation likelihood, while larger rungs do not convert added parameters into better compression. These results establish a compact continuous-input, categorical-output formulation for decoder-only financial sequence modeling.

\section{Introduction}

Decoder-only Transformers have become a general architecture for conditional sequence modeling. Their original success in natural language came from a simple objective: given a historical sequence, predict the distribution of the next token. This principle is not limited to language. Time series, market data and multivariate event streams can also be viewed as sequences in which the central modeling problem is to estimate the conditional distribution of the next event.

Financial markets, however, present a harder version of this problem. Text already comes with a discrete symbolic substrate. Financial observations do not. A market event is usually observed as a heterogeneous vector containing return magnitude, gap behavior, volatility state, calendar position and data-quality masks. These variables are noisy, heavy-tailed, regime-dependent and only partially comparable across assets. A financial Transformer therefore cannot simply inherit the natural-language notion of a token. Before next-token prediction can be meaningfully applied to markets, one must decide what the financial event is, how it should enter the model and what probability distribution the model should predict.

This paper studies the following scientific problem: \textbf{can a decoder-only Transformer learn useful probabilistic structure in financial return sequences, and which representation, supervision and architecture choices make this possible?} More concretely, we study next-period normalized return-bucket prediction for 1H FX market data. The target is not a raw price level or a scalar point forecast. The target is a full probability distribution over the next volatility-normalized return bucket. The main metric is negative log-likelihood (NLL) in bits per return event.

Return predictability is weak, non-stationary and easy to overstate. A model can appear to learn from financial data while only reproducing unconditional class imbalance, first-order transition persistence or shallow engineered-feature rules. The primary test is therefore whether the model improves held-out return likelihood beyond train-fitted probabilistic baselines, including Frequency, Markov and LightGBM. Without this baseline-compression test, representation and architecture ablations would compare variants of a modeling setup whose predictive value had not been established.

The problem is also difficult because financial data create a representation dilemma. Discretizing all market variables before they enter the Transformer makes the task compatible with categorical next-token training, but it destroys local numerical geometry. Two returns, gaps or volatility states that fall into the same bucket become identical to the input encoder, even when their relative magnitudes may carry useful information. This is costly in finance, where useful signals are often weak, conditional and concentrated in small distributional differences.

The opposite approach also has limitations. Keeping the entire task continuous often forces the researcher into regression losses, quantile losses or parametric density assumptions. These objectives can be useful, but they do not naturally preserve the language-model-style likelihood discipline of predicting a full categorical distribution over future states. Financial returns are not well summarized by a single point estimate. A distributional target is more appropriate because the model must represent center mass, skew, tail probability and regime-dependent uncertainty.

VAIOM resolves this tension by separating the input representation problem from the output likelihood problem. The model receives continuous multivariate financial event vectors as input, preserving the numerical geometry of market variables. At the output, it predicts a categorical distribution over the next normalized return bucket, preserving cross-entropy training, bits-per-event evaluation and direct comparison with probabilistic baselines. The formulation retains next-token-style supervision without tokenizing every input field.

The resulting architecture is a decoder-only Transformer for continuous-input, discrete-output financial sequence modeling. Each historical bar is represented as a past-only continuous market-event vector containing standardized return, gap, volatility-regime, calendar and mask information. The selected model further augments this representation with categorical asset, asset-class and timeframe metadata. These event representations are projected into the Transformer hidden space and processed causally over market histories. The primary target is the next-period normalized return bucket. Bucket boundaries, categorical maps and other fitted preprocessing parameters are estimated only on Train; fixed recursive rules such as exponentially weighted moving average (EWMA) spans retain their Train-defined specification while their states update causally from past observations. Model selection and ablation decisions use Validation, and final performance is reported on the held-out 2025 Test interval.

The empirical study addresses four research questions: whether VAIOM improves held-out next-return likelihood beyond statistical and tree-based baselines; whether continuous financial-event input improves over a discrete-token input branch under the same categorical output objective; whether full-sequence autoregressive supervision improves over last-position training; and whether auxiliary objectives and the Mixture-of-Market-States (MoMS) return head improve the main return likelihood. A separate supporting analysis evaluates how the fixed corpus supports predefined 0.9M, 5M and 15M architecture rungs.

The contribution is threefold. First, to our knowledge, VAIOM is the first decoder-only model to predict a categorical distribution over the next return bucket from continuous multivariate market-event inputs. Second, the selected Hybrid Continuous Input (HybridContIn) configuration improves Test likelihood beyond Frequency, Markov and LightGBM across three training seeds and both 2025 Test halves. Third, the ablations identify continuous event representation, full-sequence supervision, auxiliary representation shaping and mixture-structured return heads as effective design choices under the evaluated protocols. The supporting capacity experiment further shows that the 0.9M rung achieves the strongest Validation likelihood on the present corpus, without treating this local result as a universal scaling threshold.

\section{Related Work}

\subsection{Transformer-Based Time-Series Forecasting}

Transformer architectures were originally introduced for sequence transduction in natural language processing, replacing recurrent and convolutional sequence models with self-attention \citep{vaswani2017attention}. Their success motivated a large body of work adapting attention-based models to time-series forecasting. Early long-sequence forecasting models focused on the computational and memory limitations of applying vanilla self-attention to long temporal contexts. Informer introduced ProbSparse attention and a generative-style decoder for long sequence time-series forecasting \citep{zhou2021informer}. Autoformer incorporated series decomposition into the model architecture and replaced point-wise attention with an auto-correlation mechanism for long-term forecasting \citep{wu2021autoformer}. FEDformer further combined decomposition with frequency-domain modeling to capture global structure in long-horizon series \citep{zhou2022fedformer}.

Subsequent work showed that the effectiveness of Transformers in time series depends heavily on representation design. DLinear challenged the assumption that increasingly complex Transformer models are always superior, showing that simple linear models can outperform several Transformer-based forecasters on long-term forecasting benchmarks \citep{zeng2023transformers}. PatchTST reframed time-series tokens as subseries-level patches and used channel-independent Transformer processing. This design improves long-term forecasting and self-supervised representation learning \citep{nie2023patchtst}. iTransformer inverted the usual temporal-token representation by treating variates as tokens, arguing that timestamp-level tokens can fuse heterogeneous variables in a way that weakens variate-centric representation learning \citep{liu2024itransformer}. TimesNet transformed one-dimensional time series into two-dimensional tensors to model intra-period and inter-period variation \citep{wu2023timesnet}. TimeMixer later emphasized decomposable multiscale mixing across temporal resolutions \citep{wang2024timemixer}.

This literature establishes that time-series Transformers require domain-specific representation choices. VAIOM follows this lesson but studies a different task. Most of the above models are evaluated on generic forecasting benchmarks using continuous forecasting losses such as mean squared error (MSE), mean absolute error (MAE) or probabilistic forecasting metrics. VAIOM instead focuses on financial next-return distribution modeling, where the target is a discrete probability distribution over volatility-normalized return buckets and the main metric is return NLL in bits per event.

\subsection{Time-Series Foundation Models and Tokenization}

Recent time-series foundation models (TSFMs) extend the pretraining paradigm to diverse temporal datasets. Chronos is especially relevant because it explicitly treats time series as a language-like modeling problem: values are scaled, quantized into a fixed vocabulary and modeled by Transformer language-model architectures using cross-entropy loss \citep{ansari2024chronos}. TimesFM proposes a decoder-only foundation model pretrained on a large time-series corpus for zero-shot forecasting across domains and granularities \citep{das2024timesfm}. Moirai presents a masked encoder-based universal forecasting Transformer trained on the Large-scale Open Time Series Archive (LOTSA), which spans multiple domains, and targets cross-frequency, arbitrary-variate forecasting \citep{woo2024moirai}. MOMENT develops an open family of TSFMs and emphasizes the difficulty of pretraining across heterogeneous public time-series datasets \citep{goswami2024moment}. Lag-Llama proposes a decoder-only probabilistic TSFM using lags as covariates \citep{rasul2023lagllama}. Time-MoE scales TSFMs with sparse mixture-of-experts and reports billion-scale decoder-only autoregressive models trained on Time-300B \citep{shi2024timemoe}.

Another line adapts pretrained large language models (LLMs) or vision models to time series rather than training time-series-native models from scratch. One Fits All uses frozen pretrained language or vision Transformer backbones for general time-series analysis \citep{zhou2023onefitsall}. Time-LLM reprograms time-series inputs into representations suitable for a frozen LLM and adds prompt-as-prefix information for forecasting \citep{jin2024timellm}. These approaches show that time-series modeling can borrow from language-model architectures, but they do not directly resolve the representation problem for financial market events.

VAIOM differs from both tokenized and continuous forecasting foundation-model routes. Chronos-style models tokenize the input series before modeling; many forecasting foundation models retain continuous outputs or predict future trajectories. VAIOM separates the two sides of the problem: historical market events enter the model as continuous financial event vectors, while the output remains a categorical return-bucket distribution. This allows the model to preserve input-side numerical geometry while retaining language-model-style likelihood training and direct comparison against probabilistic baselines.

\subsection{Financial Market Sequence Modeling and Financial Foundation Models}

Recent work has begun to model financial numerical sequences directly with decoder-only Transformers. StockGPT discretizes daily U.S. equity returns into a vocabulary of return tokens and trains a vanilla decoder-only Transformer through autoregressive cross-entropy to predict the distribution of the next return token \citep{mai2024stockgpt}. Its formulation establishes an especially close precedent for probabilistic next-return modeling: both StockGPT and VAIOM use causal decoder-only backbones and categorical distributions over future return states. The principal representation difference is that StockGPT uses discretized scalar returns on both the input and output sides, whereas VAIOM preserves a continuous multivariate financial-event vector at the input and discretizes only the volatility-normalized return target.

Other financial sequence models operate at broader market-event or forecasting scales. Kronos discretizes multivariate candlestick information into token sequences and pretrains an autoregressive Transformer on a large multi-market corpus \citep{shi2025kronos}. MarS models order-level financial events generatively for interactive market simulation \citep{li2024mars}, while MarketGPT uses a Generative Pre-trained Transformer (GPT)-style architecture to generate limit order book (LOB) event streams inside a discrete-event simulator \citep{wheeler2024marketgpt}. FinCast takes a different continuous forecasting route: it is a large decoder-only financial foundation model that projects normalized time-series patches into continuous latent representations and produces multi-horizon point and quantile forecasts through a point-quantile objective \citep{zhu2025fincast}.

These studies establish complementary routes for decoder-only financial modeling. StockGPT retains discrete representations on both sides of the model; Kronos and order-flow models construct discrete financial-event vocabularies; and FinCast retains continuous patch representations together with continuous forecasting outputs. VAIOM studies the remaining hybrid formulation: continuous multivariate market-event input with a discrete volatility-normalized next-return likelihood. Its empirical scope is narrower and more controlled than foundation-model or simulation work. Rather than evaluating zero-shot multi-domain forecasting, synthetic market generation or downstream portfolio returns, VAIOM asks whether this representation improves held-out next-return NLL beyond train-fitted statistical and tree-based baselines.

Recent empirical work also suggests that financial time-series foundation modeling requires domain-specific care. Studies of TSFMs in finance find that off-the-shelf general TSFMs can underperform in zero-shot or fine-tuning settings, while finance-specific pretraining can provide stronger forecasting and economic results \citep{rahimikia2025revisiting}. Survey work on financial TSFMs emphasizes that financial forecasting requires attention to heavy tails, non-stationarity, evaluation protocols and risk-aware probabilistic scoring \citep{lu2025financeTSFM}. These observations motivate VAIOM's design: the model is trained and evaluated directly on financial return-distribution likelihood rather than treated as a generic forecasting benchmark.

\subsection{Machine Learning for Return Prediction and Baselines}

A separate but important literature studies machine learning for asset pricing and return prediction. Gu, Kelly and Xiu compare machine learning methods for empirical asset pricing and show that trees and neural networks can capture nonlinear predictor interactions missed by more restrictive models \citep{gu2020empirical}. This line of work motivates the use of nonlinear tabular models as serious financial baselines rather than relying only on unconditional or linear comparisons. Earlier deep learning work, such as Fischer and Krauss, applied long short-term memory (LSTM) networks to financial market prediction and compared recurrent neural models against classical machine-learning baselines \citep{fischer2018deep}. More recent work has applied Transformer-style architectures to stock return prediction, including studies that adapt forecasting Transformers to equity excess-return prediction tasks \citep{wang2025transformers}.

VAIOM differs from this return-prediction literature in both target and evaluation. Many financial machine-learning studies focus on point forecasts, directional movement, excess-return prediction or downstream portfolio performance. VAIOM focuses on probabilistic next-return bucket modeling. The primary object is not a trading rule, a point forecast or a portfolio return. It is the full conditional distribution over the next normalized return bucket, evaluated by held-out likelihood.

This distinction is why the baseline design is central. VAIOM is evaluated against Frequency, Markov and LightGBM baselines. Frequency and Markov test whether the model goes beyond unconditional class imbalance and first-order bucket persistence. LightGBM provides a tree-based nonlinear feature baseline, following the broader empirical asset-pricing lesson that nonlinear tabular models are strong competitors in financial prediction. LightGBM itself is a gradient boosting decision tree (GBDT) system designed for efficient large-scale training using techniques such as gradient-based one-side sampling and exclusive feature bundling \citep{ke2017lightgbm}. By requiring baselines to output probability distributions over return buckets, VAIOM keeps the comparison on the same probabilistic likelihood metric.

\subsection{Representation, Supervision and Output-Head Design}

VAIOM's architecture choices also connect to broader machine-learning ideas. Auxiliary objectives are closely related to multi-task learning and deep supervision. Multi-task learning uses shared representations across related tasks and can improve data efficiency or reduce overfitting, but task choice and loss balancing are nontrivial \citep{ruder2017multitask,crawshaw2020multitask}. Deeply supervised networks use auxiliary losses to make hidden-layer learning more direct \citep{lee2015deeply}. VAIOM uses Gap, VolReg and Ordinal heads in this spirit: auxiliary losses are not alternative success metrics, but representation-shaping signals judged only by whether they improve return likelihood.

The ordinal auxiliary head is motivated by the ordered nature of return buckets. Standard multi-class cross-entropy treats all wrong classes as unordered categories, while return buckets have a natural rank structure. Ordinal regression methods such as Consistent Rank Logits (CORAL) explicitly model ordered class labels and enforce rank-consistent ordinal predictions \citep{cao2020coral}. VAIOM does not replace the main categorical likelihood with an ordinal objective. Instead, it adds ordinal supervision as an auxiliary training signal while retaining return NLL as the evaluation objective.

MoMS is related to mixture-output modeling. Mixture density networks combine neural networks with mixture models to represent conditional probability distributions rather than single deterministic predictions \citep{bishop1994mixture}. VAIOM applies a categorical analogue of this idea: the return distribution is represented as a mixture of latent predictive components over return buckets. The model does not assume that these latent components are economically identified regimes. Their value is empirical and likelihood-based: MoMS is retained only if it improves return NLL relative to an independent categorical head.

Finally, VAIOM's full-sequence supervision follows the autoregressive training logic of causal sequence models. Standard causal language modeling trains a model to predict future tokens under a left-context constraint. VAIOM transfers this principle to financial event sequences by comparing last-position loss with full-sequence next-return supervision. The contribution is not the invention of full-sequence autoregressive training, but its validation as a supervision-density choice for noisy, overlapping financial return windows.

\subsection{Positioning of VAIOM}

Decoder-only financial models now cover several representation choices. StockGPT uses discrete return tokens at both input and output; FinCast maps continuous patches to continuous point and quantile forecasts; Kronos tokenizes multivariate candlestick sequences; and VAIOM combines continuous multivariate market-event input with a categorical next-return output \citep{mai2024stockgpt,zhu2025fincast,shi2025kronos}.

To our knowledge, VAIOM is the first decoder-only model to predict a categorical distribution over the next return bucket from continuous multivariate market-event inputs. Its scope is controlled held-out likelihood rather than broad foundation-model transfer, event-stream simulation or trading performance. The empirical study compares input representation, supervision density, auxiliary representation shaping, mixture-structured output heads and corpus-matched capacity within one small decoder-only model family.

\section{Methods}

\subsection{Task Definition: Next-Return Distribution Modeling}

Given a historical sequence of market events available up to time \(t\), the model predicts the probability distribution of the next normalized return bucket \(y_{a,t+1}\), where \(a\) indexes the asset. Let \(C_{a,t}\) denote the close price of asset \(a\) at time \(t\). The next-period log return, its volatility-normalized form and the bucketized target are

\[
r_{a,t+1}=\log\frac{C_{a,t+1}}{C_{a,t}},
\qquad
z_{a,t+1}=\frac{r_{a,t+1}}{\sigma_{a,t}},
\qquad
y_{a,t+1}=B(z_{a,t+1}),
\]

where \(\sigma_{a,t}\) is an exponentially weighted moving-average scale of past clean log returns through time \(t\) (short span \(=20\)), and \(B(\cdot)\) is the bucket map defined in Section~\ref{sec:buckets}.

Let \((\mathcal{F}_t)_{t\ge 0}\) be the natural filtration generated by the observed market history. The volatility estimate \(\sigma_{a,t}\) is \(\mathcal{F}_t\)-measurable: it uses only information available through time \(t\) and never uses \(r_{a,t+1}\), \(C_{a,t+1}\), or any other quantity from the prediction interval. Each input field \(x_{a,s}\) is \(\mathcal{F}_t\)-measurable for all \(s\le t\). The target \(y_{a,t+1}\) is \(\mathcal{F}_{t+1}\)-measurable and is not generally \(\mathcal{F}_t\)-measurable.

The model therefore learns

\[
P_\theta(y_{a,t+1}\mid x_{a,\le t}).
\]

The task is not raw price-level prediction, raw return regression, or directional classification. It is probabilistic next-return bucket prediction. The main evaluation metric is return negative log-likelihood, reported in bits per return event.

\subsection{Corpus Construction and Split Protocol}

The main corpus used in this paper is the 1H FX V0.2 corpus with expanded calendar fields. The corpus is constructed as a set of per-asset time series. Training samples are drawn from valid causal context windows, and no cross-asset sequence concatenation is used.

The split protocol is chronological and assigns each prediction event by the timestamp of its target bar:

\[
Train: \quad Datetime < 2024\text{-}01\text{-}01
\]

\[
Validation: \quad 2024\text{-}07\text{-}01 \leq Datetime < 2025\text{-}01\text{-}01
\]

\[
Test: \quad 2025\text{-}01\text{-}01 \leq Datetime < 2026\text{-}01\text{-}01.
\]

The interval \(2024\text{-}01\text{-}01 \leq Datetime < 2024\text{-}07\text{-}01\) is a non-supervised temporal buffer. Its target events are not used for parameter fitting, checkpoint selection, ablation comparison or reported evaluation. Its observed bars are retained only as past-only causal history: they update recursive feature states and may enter the left context of early Validation windows. This preserves deployment-time continuity without turning 2024H1 into a fitted or scored split.

No random split is used. Train determines all fitted preprocessing quantities, including bucket edges, categorical maps and serialized tokenizer state, together with fixed recursive specifications such as EWMA spans and floors. Validation and Test are transform-only in the parameter-estimation sense: no preprocessing parameter, baseline distribution or model parameter is refit. Recursive EWMA states nevertheless continue to update through the buffer, Validation and Test using only observations available by each feature's information time and strictly before its prediction target.

The paper uses the split names as follows. Train is used for fitting tokenizers, model parameters and train-fitted baselines. Validation is used for checkpoint selection, ablation comparison and model-design diagnosis. Test is the 2025 calendar-year interval and is used only for final held-out reporting of the selected main model. Data at or after \(2026\text{-}01\text{-}01\) are excluded from every result in this paper and reserved as a future holdout.

\paragraph{Test purity.} The Test interval is reported as two pre-defined halves:

\[
\begin{aligned}
\mathrm{2025H1}:&\quad 2025\text{-}01\text{-}01 \leq Datetime < 2025\text{-}07\text{-}01,\\
\mathrm{2025H2}:&\quad 2025\text{-}07\text{-}01 \leq Datetime < 2026\text{-}01\text{-}01.
\end{aligned}
\]

Evaluation was performed separately for each half after the main model was fully selected. No Test statistic, Test likelihood, Test comparison or Test visualization entered any design decision, ablation choice, checkpoint selection, hyperparameter tuning or model-selection step. The main model was selected on Validation alone; only then was it evaluated on Test.

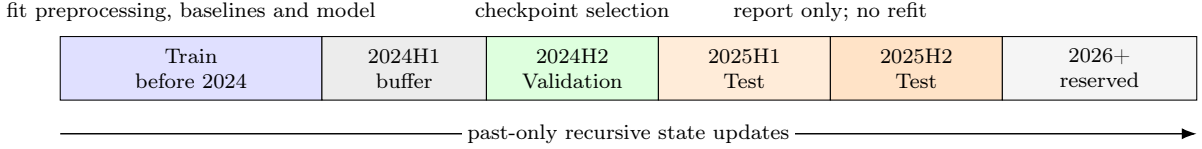
\begin{figure}[t]
\centering
\resizebox{\textwidth}{!}{%
\begin{tikzpicture}[font=\scriptsize, x=1cm, y=1cm]
\def\h{0.85}
\draw[fill=blue!12] (0,0) rectangle (3.5,\h);
\draw[fill=gray!15] (3.5,0) rectangle (5.7,\h);
\draw[fill=green!13] (5.7,0) rectangle (8.0,\h);
\draw[fill=orange!15] (8.0,0) rectangle (10.3,\h);
\draw[fill=orange!22] (10.3,0) rectangle (12.6,\h);
\draw[fill=gray!8] (12.6,0) rectangle (15.2,\h);
\node[align=center] at (1.75,0.43) {Train\\before 2024};
\node[align=center] at (4.6,0.43) {2024H1\\buffer};
\node[align=center] at (6.85,0.43) {2024H2\\Validation};
\node[align=center] at (9.15,0.43) {2025H1\\Test};
\node[align=center] at (11.45,0.43) {2025H2\\Test};
\node[align=center] at (13.9,0.43) {2026+\\reserved};
\draw[-{Latex[length=2mm]},line width=0.55pt] (0,-0.45) -- (15.2,-0.45);
\node[align=center,fill=white,inner sep=2pt] at (7.6,-0.45) {past-only recursive state updates};
\node[align=center] at (1.75,1.18) {fit preprocessing, baselines and model};
\node[align=center] at (6.85,1.18) {checkpoint selection};
\node[align=center] at (10.3,1.18) {report only; no refit};
\end{tikzpicture}
}
\caption{Chronological split and causal information flow. Train determines fitted quantities. Recursive states update from past observations across later intervals, but no fitting occurs after Train.}
\label{fig:split_timeline}
\end{figure}

Figure~\ref{fig:split_timeline} summarizes the split and information-flow rules. Recursive feature states continue causally across boundaries, but fitting ends with Train; the buffer is unscored, Validation selects the model and 2026 onward remains untouched.

\subsection{Market Data, Price Convention and Bar Construction}

The 1H corpus is derived from one-minute FX and spot-metal open-high-low-close-volume (OHLCV) data obtained from Dukascopy. The universe contains AUDUSD, EURUSD, GBPUSD, NZDUSD, USDCAD, USDCNY, USDHKD, USDJPY, USDSGD, XAGUSD and XAUUSD. The first nine series are assigned to the FX class and XAGUSD and XAUUSD to the spot-metal class. Each asset is processed as a separate time series; windows never cross asset boundaries.

The archive exposes one open-high-low-close (OHLC) price stream rather than separate bid and ask series. No midpoint is reconstructed. In this paper, \(O_{a,t}\) and \(C_{a,t}\) therefore denote the stored open and close of that source stream; in particular, \(C_{a,t}\) is not a separately calculated \((\mathrm{bid}+\mathrm{ask})/2\) midpoint. Raw trading volume is disabled because the available FX activity field is a broker-tick proxy rather than comparable exchange volume.

Appendix Table~\ref{tab:asset_coverage} reports archived coverage and valid Train-window counts by asset. The counts sum to 1,266,989 unique valid Train windows. Bars from 2026 remain in the archive but are excluded from every result under the future-holdout rule.

One-hour bars are formed independently by symbol from fixed one-hour clock bins. Within each non-empty bin, Open is the earliest one-minute Open, High is the maximum High, Low is the minimum Low, Close is the latest one-minute Close and Volume is the sum of available one-minute volume values. The stored \texttt{Datetime} is the left boundary of the one-hour bin. Timestamps are timezone-naive in the archived parquet files and are consumed without localization or timezone conversion; split cutoffs and calendar covariates use this stored provider clock.

Empty clock bins are not synthesized. The pipeline performs no interpolation, forward filling or weekend filling, and therefore inserts no artificial Saturday or Sunday bars. Consecutive observed rows remain consecutive in the per-asset sequence across holidays and weekends. The opening displacement after any such interval is retained through

\[
g_{a,t}=\log\frac{O_{a,t}}{C_{a,t-1}},
\]

and is normalized by a past-only gap scale. Thus a weekend or market-closure jump is represented as a gap, while the missing wall-clock bins themselves do not become model events.

\paragraph{Data-quality masks.} A row is \texttt{missing} when Open or Close is null or non-positive. It is \texttt{stale} when a previous close exists and the close-to-close log return is exactly zero; the configured tolerance is \(0\). It is \texttt{bad-data} when \(|\log(C_t/C_{t-1})|>1\) or \(|\log(O_t/C_{t-1})|>1\). It is \texttt{insufficient-history} when fewer than 20 preceding clean returns are available, and \texttt{scale-zero} when the past-only volatility scale is at or below \(10^{-8}\). The row-level \texttt{mask-any} flag is the logical OR of these five masks. Masked categorical input fields receive index \(0\); ContIn retains the numerical mask channels. A next-return target is invalid when the next close-to-close return is non-finite, exactly zero, exceeds the absolute log-return threshold, or the next row has missing/non-positive Open or Close. Invalid targets receive index \(0\) and never enter any loss or reported likelihood denominator.

\paragraph{Continuous dynamic features.} Define the close return and opening gap as

\[
r_{a,t}=\log\frac{C_{a,t}}{C_{a,t-1}},
\qquad
g_{a,t}=\log\frac{O_{a,t}}{C_{a,t-1}}.
\]

Let \(\widetilde r_{a,t}\) equal \(r_{a,t}\) unless the row is missing, stale or has \(|r_{a,t}|>1\), in which case it is missing. Likewise, \(\widetilde g_{a,t}\) equals \(g_{a,t}\) unless the row is missing or has \(|g_{a,t}|>1\). For a possibly missing series \(q\), define

\[
\mathcal E_s(q_{\le t})
=
\operatorname{EWMMean}
\!\left(q_{\le t};\ \alpha=\frac{2}{s+1},\
\ \texttt{adjust=False},\ \texttt{min\_periods=1}\right),
\]

using the implementation's default missing-value handling, and

\[
\sigma^{(s)}_{a,t}
=
\max\!\left(10^{-8},\sqrt{\mathcal E_s(\widetilde r^{,2}_{a,\le t})}\right),
\qquad
\sigma^{\mathrm{gap}}_{a,t}
=
\max\!\left(10^{-8},\sqrt{\mathcal E_{20}(\widetilde g^{,2}_{a,\le t})}\right).
\]

The four dynamic ContIn fields are therefore

\[
\operatorname{ret\_z}_{a,t}
=\frac{r_{a,t}}{\sigma^{(20)}_{a,t-1}},
\qquad
\operatorname{gap\_z}_{a,t}
=\frac{g_{a,t}}{\sigma^{\mathrm{gap}}_{a,t-1}},
\]

\[
\operatorname{relative\_log\_vol}_{a,t}
=\log\sigma^{(20)}_{a,t}-\log\sigma^{(120)}_{a,t},
\qquad
\operatorname{sigma\_through\_t}_{a,t}
=\sigma^{(20)}_{a,t}.
\]

Thus \texttt{sigma\_through\_t} is the untransformed short-span EWMA scale through the current row. Non-finite continuous fields are serialized as zero and all 25 numerical inputs are clipped to \([-32,32]\) only when they enter the model. The Gap bucket map uses the same fixed tail cutoffs and Train-fitted interior-quantile procedure as the return map; its exact serialized 1H edges are reported in Appendix~\ref{app:bucket_validity}.

\subsection{Return Bucket Construction}
\label{sec:buckets}

The output vocabulary is defined over normalized return buckets. The normalized return \(z_{a,t+1}\) removes raw price level and asset-scale effects by measuring each return in units of the asset's own past volatility. Bucket boundaries are fitted on Train and then fixed for the buffer, Validation and Test.

For a series of clean returns \(\tilde r\), the implemented scale is

\[
\sigma^{(s)}_{a,t}
=
\max\!\left(10^{-8},
\sqrt{\operatorname{EWMA}_{s}\!\left(\tilde r_{a,\leq t}^{\,2}\right)}\right),
\]

where \texttt{EWMA} is the pandas exponentially weighted mean with \texttt{span}=\(s\), \texttt{adjust=False} and \texttt{min\_periods}=1. Equivalently, on consecutive non-missing observations its decay coefficient is \(\alpha=2/(s+1)\). The spans, floor and update rule are fixed before out-of-sample (OOS) evaluation; they are not re-estimated on Validation or Test. The EWMA state itself is dynamic and continues recursively through the buffer, Validation and Test using past-only observations. The target normalization uses \(s=20\) and the scale through time \(t\), so \(z_{a,t+1}=r_{a,t+1}/\sigma^{(20)}_{a,t}\). The current-return ContIn field uses the analogous scale computed through \(t-1\). The long volatility input uses \(s=120\).

Let the bucket edges be

\[
-\infty=b_0<b_1<\cdots<b_{K-1}<b_K=+\infty,
\]

where \(K=16\) is the number of return buckets. The bucket map is

\[
B(z)=j
\quad\Longleftrightarrow\quad
b_{j-1}<z\le b_j,
\qquad j\in\{1,\ldots,K\}.
\]

The edges are hybrid: fixed tail cutoffs \((-8,-5,-3,-2,+2,+3,+5,+8)\) in units of \(\sigma_{a,t}\) and seven interior quantile edges \((0.1,0.2,0.35,0.5,0.65,0.8,0.9)\) fitted on the Train distribution of \(z\) values inside \((-2,+2)\). The special index \(0\) marks missing or invalid observations.

For the reported 1H V0.2 corpus, the fitted return edges, shown to four decimal places, are

\[
(-\infty,-8,-5,-3,-2,-1.0085,-0.6241,-0.2697,
0.0118,0.2833,0.6362,1.0206,2,3,5,8,+\infty).
\]

These 17 edges define exactly 16 non-missing target buckets; the serialized tokenizer state, rather than a refit, is used by every Validation and Test run.

\subsection{Input Representations: DiscIn, ContIn and HybridContIn}

The primary representation study compares Discrete Input (DiscIn) and pure Continuous Input (ContIn) while keeping the output target fixed; Hybrid Continuous Input (HybridContIn) is evaluated afterward as a refinement of the continuous branch.

In \textbf{DiscIn}, raw-volume field \texttt{volume\_bucket} is disabled. The variant contains the 15 categorical fields in Table~\ref{tab:discin_fields}. Return, gap and volatility-regime bucket maps are fitted on Train, with index \(0\) marking missing values. Asset identifiers are assigned in sorted-symbol order: AUDUSD through XAUUSD receive integer values 1 through 11, and \(0\) is reserved for an unknown asset. The Train-fitted class map is FX \(\mapsto 0\) and spot metal (internal label \texttt{SPOT}) \(\mapsto 1\); the sole timeframe maps 1H \(\mapsto 0\). Calendar fields are extracted from the stored provider-clock timestamp. The field \texttt{years\_since\_2000} is clipped to \([0,63]\). Month, day and day-of-year use one-based integers; day-of-week, hour, minute and second use zero-based integers.

The complete DiscIn schema is listed in Appendix Table~\ref{tab:discin_fields}.

The \textbf{ContIn} variant contains exactly the 25 numerical fields in Table~\ref{tab:contin_fields}. Raw-volume field \texttt{volume\_z} is disabled. Non-finite values are replaced by \(0\), and every numerical field is clipped elementwise to \([-32,32]\) before input projection.

The complete numerical schema is listed in Appendix Table~\ref{tab:contin_fields}; metadata fields are injected separately.

For the continuous calendar fields, \texttt{years\_since\_2000\_norm} equals the clipped year offset divided by 63. For each periodic calendar variable \(v\) with period \(P\), the two channels are \(\sin(2\pi(v-o)/P)\) and \(\cos(2\pi(v-o)/P)\), where \(o=1\) for month, day-of-month and day-of-year and \(o=0\) for day-of-week, hour, minute and second. The corresponding periods are 12, 31, 366, 7, 24, 60 and 60.

After completing the primary DiscIn-versus-ContIn representation study, we evaluated \textbf{HybridContIn}, a metadata-augmented extension of the continuous-input branch. The original 25-dimensional continuous market-event vector remains unchanged. Asset identity, asset class and timeframe are additionally represented through learned categorical embeddings and fused with the continuous event projection before the Transformer backbone.

HybridContIn is treated as a subsequent model refinement rather than as part of the primary representation ablation. The DiscIn-versus-pure-ContIn comparison tests whether continuous dynamic market variables improve return likelihood relative to tokenized market variables under the same output target and return-likelihood objective. The HybridContIn experiment then tests whether static categorical identity information provides complementary predictive information after the continuous representation has been established. The selected main model uses the HybridContIn extension.

Raw trading volume is disabled in these experiments. \textbf{VolReg} always denotes volatility-regime information and supervision.

\begin{figure}[t]
\centering
\begin{tikzpicture}[
font=\footnotesize,
>={Latex[length=2mm]},
branch/.style={draw, rounded corners=1.5pt, fill=gray!7, align=center, text width=4.15cm, minimum height=1.65cm, inner sep=5pt},
shared/.style={draw, rounded corners=1.5pt, fill=blue!5, align=center, text width=10.0cm, minimum height=0.78cm, inner sep=5pt},
head/.style={draw, rounded corners=1.5pt, fill=gray!4, align=center, text width=4.8cm, minimum height=0.85cm, inner sep=5pt},
arrow/.style={->, line width=0.55pt}
]
\node[branch] (disc) at (-5.0,0) {\textbf{DiscIn}\\15 categorical fields\\independent 8-d embeddings\\concatenate 120-d $\rightarrow$ MLP $\rightarrow$ 128-d};
\node[branch] (cont) at (0,0) {\textbf{Pure ContIn}\\25 continuous fields\\MLP $25\rightarrow128\rightarrow128$\\128-d event embedding};
\node[branch] (hybrid) at (5.0,0) {\textbf{HybridContIn}\\25 fields $\rightarrow$ 128-d projection\\asset/class/timeframe: $3\times8$-d\\concatenate 152-d $\rightarrow$ fusion MLP $\rightarrow$ 128-d};

\node[shared] (pos) at (0,-2.25) {Learned absolute positional embeddings (positions 0--511)};
\node[shared] (decoder) at (0,-3.65) {Four-layer pre-normalized causal Transformer decoder\\4 attention heads; hidden width 128; feed-forward width 512};
\node[head] (return) at (-2.75,-5.20) {\textbf{Return-distribution head}\\Independent categorical or MoMS};
\node[head] (aux) at (2.75,-5.20) {\textbf{Auxiliary heads}\\Gap, VolReg and Ordinal};
\node[shared] (scope) at (0,-6.65) {Last-position or FullSeq causal supervision};

\draw[arrow] (disc.south) -- (pos.north west);
\draw[arrow] (cont.south) -- (pos.north);
\draw[arrow] (hybrid.south) -- (pos.north east);
\draw[arrow] (pos) -- (decoder);
\draw[arrow] (decoder.south) -- ++(0,-0.30) -| (return.north);
\draw[arrow] (decoder.south) -- ++(0,-0.30) -| (aux.north);
\draw[arrow] (return.south) -- (scope.north west);
\draw[arrow] (aux.south) -- (scope.north east);
\end{tikzpicture}
\caption{VAIOM architecture. Alternative DiscIn, pure ContIn and HybridContIn encoders feed the same causal decoder, return head, auxiliary heads and supervision interface.}
\label{fig:input_architecture}
\end{figure}
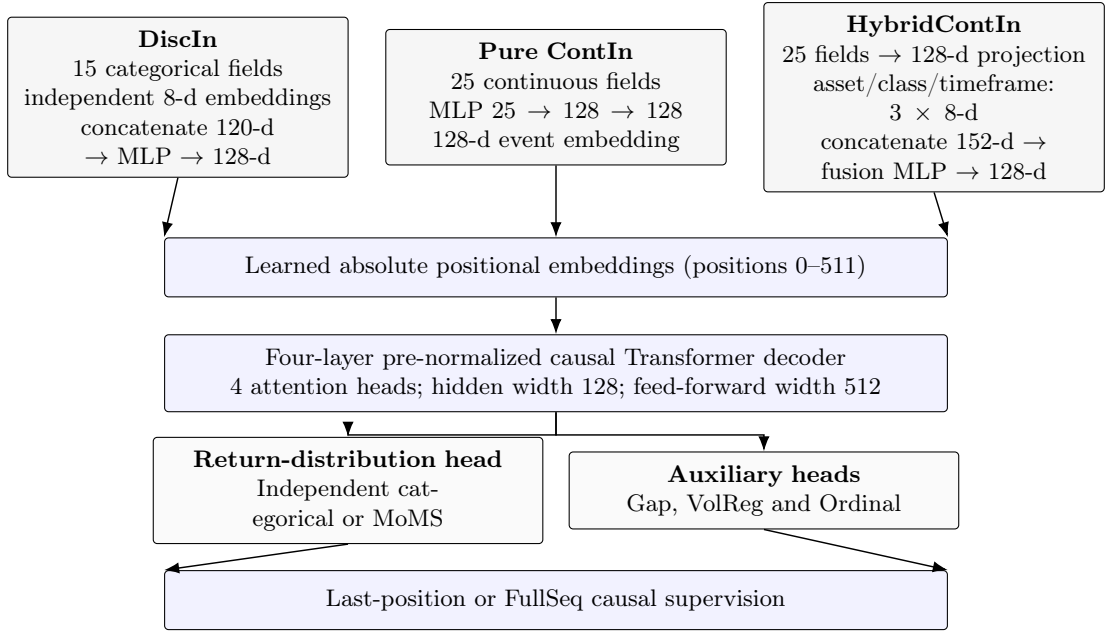

Figure~\ref{fig:input_architecture} summarizes the three alternative input branches and the shared causal backbone. DiscIn and pure ContIn define the primary representation comparison; HybridContIn augments the continuous branch with static categorical metadata.

\subsection{VAIOM Backbone}

VAIOM is a causal decoder-only Transformer over financial event embeddings, built from the self-attention architecture introduced by \citet{vaswani2017attention}. DiscIn and pure ContIn use a two-layer input multilayer perceptron (MLP) with Gaussian error linear unit (GELU) activation:

\[
e_t=W_2\,\operatorname{GELU}(W_1u_t+b_1)+b_2,
\]

where \(e_t\in\mathbb{R}^{d}\) and \(d=128\) for the main model. For pure ContIn, \(u_t\in\mathbb{R}^{25}\) is the clipped numerical event vector, so the projection is \(25\rightarrow128\rightarrow128\). For the selected HybridContIn model, the implemented fusion is concatenation followed by projection. Specifically,

\[
c_{a,t}=\operatorname{GELU}\!\left(W_c x^{\mathrm{cont}}_{a,t}+b_c\right),
\]

\[
e_{a,t}=W_2\,\operatorname{GELU}\!\left(
W_1
\left[
c_{a,t};
E_{\mathrm{asset}}(a);
E_{\mathrm{class}}(\kappa(a));
E_{\mathrm{tf}}(\tau(a))
\right]
+b_1
\right)+b_2.
\]

Here \(\kappa(a)\) denotes the asset class and \(\tau(a)\) the timeframe. The continuous projection \(c_{a,t}\in\mathbb{R}^{128}\); each metadata embedding is 8-dimensional; the concatenated vector is therefore 152-dimensional; and the fusion MLP has shape \(152\rightarrow128\rightarrow128\) with GELU activation. This matches the implemented order: continuous projection, embedding lookup, concatenation and fused projection. For DiscIn, each of the 15 fields has an independent learned embedding table. The per-field embedding width is

\[
d_f=\max\!\left(8,\left\lfloor\frac{d}{15}\right\rfloor\right)=8.
\]

The 15 field embeddings are concatenated into \(u_t\in\mathbb{R}^{120}\) and passed through the same MLP form, \(120\rightarrow128\rightarrow128\). Field embeddings are concatenated, not summed or averaged.

A learned absolute positional embedding \(p_s\in\mathbb{R}^{128}\), indexed by sequence position \(s\in\{0,\ldots,511\}\), is added to each projected event: \(h_s^{(0)}=e_s+p_s\). No sinusoidal or rotary positional encoding is used.

Each Transformer block is pre-normalized. Layer normalization (LayerNorm) is applied before four-head causal self-attention, followed by a residual addition; a second LayerNorm precedes the feed-forward network, followed by another residual addition. The feed-forward network has inner dimension \(4d=512\), GELU activation and shape \(128\rightarrow512\rightarrow128\). Attention dropout and feed-forward output dropout are both \(0.1\). A final LayerNorm is applied after the fourth block and before all return and auxiliary heads. The causal mask is strictly upper triangular, so position \(s\) cannot attend to positions later than \(s\).

\subsection{Output Heads: Independent Categorical Head and MoMS}

The simplest output head is an independent categorical return head. Let \(h\in\mathbb{R}^{d_{\mathrm{emb}}}\) be the Transformer hidden state at the prediction position, \(W\in\mathbb{R}^{K\times d_{\mathrm{emb}}}\) and \(b\in\mathbb{R}^{K}\). The probability of bucket \(j\) is

\[
p_j(h)
=
P(y=j\mid h)
=
\frac{\exp(w_j^\top h+b_j)}
{\sum_{\ell=1}^{K}\exp(w_\ell^\top h+b_\ell)},
\]

or, in vector form, \(p(h)=\mathrm{softmax}(Wh+b)\) with \(P(y=j\mid h)=p_j(h)\).

VAIOM also evaluates a MoMS head. MoMS follows the general mixture-output principle of representing a conditional predictive distribution through multiple latent components and learned mixture weights \citep{bishop1994mixture}, adapted here to categorical return-bucket distributions. MoMS parameterizes the return distribution as a mixture of \(K_{\mathrm{states}}\) latent predictive components, with \(K_{\mathrm{states}}=4\). Let \(W_\pi\in\mathbb{R}^{K_{\mathrm{states}}\times d_{\mathrm{emb}}}\), \(b_\pi\in\mathbb{R}^{K_{\mathrm{states}}}\), \(W_k\in\mathbb{R}^{K\times d_{\mathrm{emb}}}\) and \(b_k\in\mathbb{R}^{K}\). The mixture weights and component distributions are

\[
\pi_k(h)=\mathrm{softmax}(W_\pi h+b_\pi)_k,
\qquad
p_{k,j}(h)=P_k(y=j\mid h)=\mathrm{softmax}(W_k h+b_k)_j.
\]

The blended return distribution is

\[
P(y=j\mid h)=\sum_{k=1}^{K_{\mathrm{states}}}\pi_k(h)\,p_{k,j}(h),
\]

which is a valid categorical distribution because \(\sum_k\pi_k(h)=1\) and \(\sum_j p_{k,j}(h)=1\) for every \(k\).

MoMS is used only as a return-distribution output-head hypothesis. The learned mixture components are not assumed to correspond to identifiable economic regimes. Their value is judged only by whether MoMS improves return NLL relative to an independent categorical head.

\subsection{Training Objective}

The primary training objective is return NLL. Under an independent categorical head, the per-position loss is

\[
L_{\mathrm{return}}
= -\log P(y_{\mathrm{true}}\mid h)
= -\log p_{y_{\mathrm{true}}}(h).
\]

Under MoMS, the exact per-position return NLL is written in logsumexp form for numerical stability:

\[
L_{\mathrm{return}}
=
-\operatorname{logsumexp}_{k=1}^{K_{\mathrm{states}}}
\left(
\log \pi_k(h)+\log p_{k,\,y_{\mathrm{true}}}(h)
\right).
\]

VAIOM optionally adds auxiliary objectives. Through multi-task training, these objectives shape the shared backbone representation and provide deep supervision \citep{ruder2017multitask,crawshaw2020multitask,lee2015deeply}. The full per-position loss is

\[
L
= L_{\mathrm{return}}
+ \lambda_{\mathrm{gap}} L_{\mathrm{gap}}
+ \lambda_{\mathrm{volreg}} L_{\mathrm{volreg}}
+ \lambda_{\mathrm{ord}} L_{\mathrm{ord}}.
\]

The auxiliary targets are all next-period quantities, constructed only from information in the prediction interval, and are never fed back as inputs at prediction time.

\paragraph{Gap auxiliary.} The next-period gap is

\[
g_{a,t+1}=B_{\mathrm{gap}}\!\left(
\frac{\log(O_{a,t+1}/C_{a,t})}{\sigma^{\mathrm{gap}}_{a,t}}
\right),
\]

where \(O_{a,t+1}\) is the next open, \(C_{a,t}\) is the current close, \(\sigma^{\mathrm{gap}}_{a,t}\) is a past-only gap scale, and \(B_{\mathrm{gap}}\) is a signed-edge bucket map fitted on Train. The gap loss is

\[
L_{\mathrm{gap}}
= -\log P_\theta(g_{a,t+1}\mid x_{a,\le t}).
\]

\paragraph{VolReg auxiliary.} The next-period volatility-regime bucket is

\[
v_{a,t+1}=B_{\mathrm{vol}}\!\left(
\log\sigma^{\mathrm{short}}_{a,t+1}-\log\sigma^{\mathrm{long}}_{a,t+1}
\right),
\]

where \(\sigma^{\mathrm{short}}\) and \(\sigma^{\mathrm{long}}\) are short-span (20) and long-span (120) EWMA scales of past clean log returns, and \(B_{\mathrm{vol}}\) is a four-quantile edge map \((0.2,0.4,0.6,0.8)\) fitted on Train, producing five regimes. The volreg loss is

\[
L_{\mathrm{volreg}}
= -\log P_\theta(v_{a,t+1}\mid x_{a,\le t}).
\]

The label \(v_{a,t+1}\) is used only as an auxiliary training target and is never used as an input at prediction time, so there is no leakage into the return head.

\paragraph{Ordinal auxiliary.} The ordinal head outputs \(K-1\) threshold logits \(\ell=(\ell_1,\ldots,\ell_{K-1})\) via a linear projection \(W_{\mathrm{ord}}h+b_{\mathrm{ord}}\). For a target bucket \(y\in\{1,\ldots,K\}\), define the cumulative binary labels

\[
\ell^{\star}_j=\mathbf{1}[j<y],
\qquad j\in\{1,\ldots,K-1\}.
\]

The ordinal loss is the average binary cross-entropy (BCE) over the \(K-1\) thresholds:

\[
L_{\mathrm{ord}}
=
\frac{1}{K-1}
\sum_{j=1}^{K-1}
\mathrm{BCE}(\ell_j,\ell^{\star}_j),
\]

where \(\mathrm{BCE}(\ell_j,\ell^{\star}_j)= -\ell^{\star}_j\log\sigma(\ell_j)-(1-\ell^{\star}_j)\log(1-\sigma(\ell_j))\), and \(\sigma(\cdot)\) is the sigmoid function. This is an independent-binary decomposition of the ordinal structure rather than CORAL: unlike the rank-consistent cumulative formulation of \citet{cao2020coral}, each threshold classifier is parameterized independently and no shared-weight or cumulative-link constraint is imposed. The ordinal loss is used only as a training signal; evaluation always reports return NLL alone. Auxiliary-task accuracy or auxiliary loss is not used as a primary success metric.

\subsection{Supervision Scope: Last-Position and Full-Sequence}

Full-sequence supervision follows the general causal autoregressive training logic of decoder-based sequence models, in which multiple positions are trained under a left-context constraint \citep{vaswani2017attention,das2024timesfm,rasul2023lagllama}. The paper compares two supervision scopes. Consider a sampled causal context window of length \(C\) for asset \(a\), starting at index \(u\): \(X_{a,u:u+C-1}\). A position is valid when

\[
\mathrm{valid\_base}_{a,s}
=
\mathbf{1}[y_{a,s+1}>0]
\land \neg\mathrm{mask\_any}_{a,s}
\land \neg\mathrm{target\_bad}_{a,s}.
\]

Here \(y_{a,s+1}>0\) marks a serialized non-missing next-return target, \(\mathrm{mask\_any}_{a,s}\) is the current-row data-quality mask and \(\mathrm{target\_bad}_{a,s}\) is the serialized \texttt{target\_mask\_bad} flag. The implementation stores this conjunction as \texttt{valid\_base}.

Return and Ordinal use the base indicator directly. Gap and VolReg additionally require their own next-period target indices to be nonzero. With \(q\in\{r,g,v,o\}\) denoting return, Gap, VolReg and Ordinal, respectively, define

\[
\begin{aligned}
m^{(r)}_{a,s}&=\mathrm{valid\_base}_{a,s},
&m^{(o)}_{a,s}&=\mathrm{valid\_base}_{a,s},\\
m^{(g)}_{a,s}&=\mathrm{valid\_base}_{a,s}\mathbf 1[y^{(g)}_{a,s+1}>0],
&m^{(v)}_{a,s}&=\mathrm{valid\_base}_{a,s}\mathbf 1[y^{(v)}_{a,s+1}>0],
\end{aligned}
\]

and, for one window,

\[
\mathcal I_q(a,u)
=\{s:u\le s\le u+C-1,\ m^{(q)}_{a,s}=1\}.
\]

In the \textbf{last-position} setting, windows are sampled with a return-valid endpoint, so \(s^{\star}=u+C-1=\max\mathcal I_r(a,u)\). The return loss is

\[
L_{\mathrm{last}}
=
-\log P_\theta\!\left(y_{a,s^{\star}+1}\mid X_{a,u:s^{\star}}\right).
\]

In the \textbf{full-sequence} setting, reduction is global over all valid positions in the current micro-batch, not a mean of per-window means. For a micro-batch \(\mathcal B=\{(a_b,u_b)\}_{b=1}^{B}\), define

\[
\mathcal I_q(\mathcal B)
=
\{(b,s):1\le b\le B,\ u_b\le s\le u_b+C-1,\
m^{(q)}_{a_b,s}=1\}.
\]

Each task is reduced separately:

\[
L_q
=
\frac{1}{|\mathcal I_q(\mathcal B)|}
\sum_{(b,s)\in\mathcal I_q(\mathcal B)}
\ell^{(q)}_{b,s},
\qquad q\in\{r,g,v,o\},
\]

and the implemented objective is

\[
L_{\mathrm{full}}
=L_r
+\lambda_{\mathrm{gap}}L_g
+\lambda_{\mathrm{volreg}}L_v
+\lambda_{\mathrm{ord}}L_o.
\]

Equivalently, \(L_q=\sum_{b,s}m^{(q)}_{b,s}\ell^{(q)}_{b,s}/\sum_{b,s}m^{(q)}_{b,s}\). The sampler guarantees a return-valid final position in every sampled window, so \(|\mathcal I_r(\mathcal B)|>0\); Ordinal uses the same return target and mask. If a micro-batch contains no valid Gap or VolReg targets, that auxiliary term is omitted for that forward pass. Four fixed-size micro-batches are accumulated per optimizer update, with each micro-batch reduced as above.

The same causal masking rule is preserved in both scopes: each position can attend only to its past. Full-sequence training therefore increases supervised signal density without changing the prediction target. Optimizer steps are not directly comparable across the two scopes, because each sampled window supplies \(|\mathcal I_r(a,u)|\) return targets under FullSeq but only one under last-position supervision.

Operationally, FullSeq constructs task target tensors of shape \((B,C)\). Invalid return and auxiliary targets are replaced by index \(0\), and each head selects its own positive-index positions before taking its global mean. No padding, invalid target or masked current row contributes to any numerator or denominator. Causal self-attention masking remains unchanged, so logits at position \(s\) depend only on \(X_{a,u:s}\).

FullSeq is treated as a training protocol, not as a different model architecture. The final main model is the \textbf{0.9M HybridContIn MoMS Gap/VolReg/Ordinal FullSeq checkpoint selected on Validation}. This selected checkpoint is then evaluated on the held-out Test split.

\subsection{Optimization and Reproducibility Settings}

The selected main model uses Adam with decoupled weight decay (AdamW) at learning rate \(3\times10^{-4}\), PyTorch default coefficients \((\beta_1,\beta_2)=(0.9,0.999)\), \(\epsilon=10^{-8}\) and weight decay \(0.01\). The learning rate is constant: no warm-up and no learning-rate scheduler are used. Gradient norm is clipped to \(1.0\). Transformer dropout is \(0.1\).

The micro-batch contains 32 windows. Four micro-batches are accumulated before each optimizer update, giving an effective batch of 128 windows per optimizer step. Training uses 16-bit floating-point (FP16) automatic mixed precision and data parallelism. The model, NumPy sampler, central processing unit (CPU) Torch random-number generator (RNG) and all Compute Unified Device Architecture (CUDA) RNGs use seed 17. The selected main checkpoint is the 5.5K-step FullSeq checkpoint chosen by Validation return NLL; the reported Test run loads this serialized checkpoint without further fitting.

The canonical main checkpoint uses random seed 17. To test whether its baseline advantage depends on a favorable initialization or training-sample order, the selected HybridContIn configuration is subsequently retrained with seeds 29 and 43. All three runs use the same corpus, architecture, optimizer, auxiliary weights, FullSeq supervision scope, 5.5K-step training budget and evaluation protocol. No architecture, hyperparameter or checkpoint rule changes across seeds. Each checkpoint is evaluated on both 2025H1 and 2025H2.

This multi-seed experiment evaluates the training stability of the already selected configuration. Individual Test results and period-specific means and sample standard deviations are reported without selecting the best seed; seed 17 remains the canonical main checkpoint. All six seed-period comparisons use exactly the same fixed LightGBM fit, whose Test NLL is 3.2057 bits/event in 2025H1 and 3.1785 in 2025H2.

\subsection{Baselines}

The paper uses three primary probabilistic baselines. All are fit only on Train; Validation and Test are transform-only evaluation splits. All baselines output full probability distributions over return buckets rather than only argmax class predictions.

The \textbf{Frequency baseline} uses the empirical Train distribution of return buckets with Laplace (Dirichlet) smoothing. Let \(C_i\) be the Train count of bucket \(i\) among valid targets and let \(\alpha=1\). Then

\[
\hat p_i
=
P(y=i)
=
\frac{C_i+\alpha}{N_{\mathrm{Train}}+\alpha K},
\]

where \(N_{\mathrm{Train}}=\sum_j C_j\) and \(K=16\). The smoothing constant \(\alpha=1\) guarantees strictly positive probability on every bucket, so the baseline NLL is finite even if a bucket has zero Train occupancy.

The \textbf{Markov baseline} is the Train-fitted first-order conditional transition distribution with the same Laplace prior. Let \(C(i\to j)\) be the Train count of transitions from bucket \(i\) to bucket \(j\). Then

\[
P(y_{t+1}=j\mid y_t=i)
=
\frac{C(i\to j)+\alpha}
{C(i)+\alpha K},
\]

with \(\alpha=1\) and \(K=16\). The transition matrix includes an extra row for the missing-token state \(i=0\).

The \textbf{LightGBM baseline} is a tree-based probabilistic classifier implemented using the gradient-boosting framework of \citet{ke2017lightgbm} and trained on features available at prediction time. Let \(\phi(x_{a,t})\) denote the feature map. In this paper, \(\phi(x_{a,t})=x_{a,t}\): LightGBM consumes the full event vector at the last timestep of the sampled window, with no autoregressive context. For the DiscIn variant, all 15 input fields are treated as categorical. For pure ContIn, all 25 continuous fields are numerical. For the HybridContIn main-model comparison, LightGBM receives the same 25 numerical fields together with asset identity, asset class and timeframe as categorical features, giving 28 single-bar fields. The predicted distribution is

\[
\hat p^{\mathrm{LightGBM}}_j(x_{a,t})
=
P_{\mathrm{LightGBM}}(y_{a,t+1}=j\mid \phi(x_{a,t})).
\]

For numerical stability, each baseline probability vector is clipped at \(\epsilon=10^{-12}\) and then renormalized before NLL computation. For any baseline distribution \(\hat p\), define

\[
q_j=\max(\epsilon,\hat p_j),
\qquad
\tilde p_j=\frac{q_j}{\sum_{\ell=1}^{K}q_\ell}.
\]

The resulting \(\tilde p\) remains a valid categorical distribution, and the bits-per-event value is \(-\log_2 \tilde p_{y_{a,t+1}}\).

\paragraph{Baseline specification and fairness caveat.} Table~\ref{tab:baseline_spec} summarizes the comparison. VAIOM consumes a 512-bar context, whereas LightGBM receives only the current feature vector \(x_{a,t}\). LightGBM is therefore a nonlinear single-bar baseline in the spirit of \citep{gu2020empirical}, not an autoregressive sequence baseline. Both models predict the same target on the same valid positions, but the comparison measures the combined value of learned sequence representation and temporal context. Appendix~\ref{app:baseline_details} gives the complete LightGBM hyperparameters.

\begin{table}[h]
\centering
\caption{Baseline specification.}
\label{tab:baseline_spec}
\begin{tabular}{llll}
\toprule
Property & Frequency & Markov & LightGBM \\
\midrule
Input & --- & previous bucket \(y_t\) & \(x_{a,t}\) (last bar only) \\
Context & unconditional & order 1 & one bar; no lag stack \\
Features & none & \(y_t\) & 28 HybridContIn fields \\
Fit & Train counts & Train transitions & Train; pooled; capped at 1M \\
\bottomrule
\end{tabular}
\end{table}

The common fixed LightGBM is fitted once on the pre-2024 Train sample using 25 numerical fields and three categorical metadata identifiers, then applied unchanged to both 2025 Test periods. Its Train holdout NLL is 3.1656 bits/event.

\subsection{Evaluation and Ablation Protocol}

For an evaluation set \(\mathcal{D}_{\mathrm{eval}}\) containing positions that satisfy \(\mathrm{valid\_base}_{a,t}=1\), average return NLL is

\[
\bar L_{\mathrm{return}}
=
-\frac{1}{|\mathcal{D}_{\mathrm{eval}}|}
\sum_{(a,t)\in\mathcal{D}_{\mathrm{eval}}}
\log P_\theta(y_{a,t+1}\mid x_{a,\le t}).
\]

All comparisons report \(\bar L_{\mathrm{return}}/\log 2\) in bits per event; lower values are better. For main Test reporting and Validation ablations, \(\mathcal{D}_{\mathrm{eval}}\) contains the final valid position of each sampled window, so each denominator event is one held-out next-return prediction. Unless stated otherwise,

\[
\Delta=\text{model bits}-\text{baseline bits},
\]

so negative \(\Delta\) favors VAIOM.

The Test split is used only for the selected main model. All design ablations are evaluated on Validation unless explicitly stated otherwise. This separates final held-out reporting from model-design diagnosis.

The ablation protocol changes one design axis where possible:

\begin{enumerate}
    \item \textbf{Input representation:} DiscIn versus ContIn.
    \item \textbf{Supervision density:} last-position loss versus full-sequence loss.
    \item \textbf{Auxiliary objectives:} no auxiliary heads, Gap, VolReg and Ordinal variants.
    \item \textbf{Output head:} independent categorical head versus MoMS.
    \item \textbf{Architecture rung:} 0.9M, 5M and 15M complete VAIOM rungs.
\end{enumerate}

Some historical component ablations are used as diagnostic evidence when exact ContIn-matched variants are unavailable. Such cases are marked in the Results section and are not interpreted as ContIn-only causal claims.

\subsection{Paired Test Uncertainty}

Statistical uncertainty for the selected VAIOM--LightGBM Test comparison is estimated from paired event losses. Let \(\ell^{\mathrm{V}}_{a,t}\) and \(\ell^{\mathrm{L}}_{a,t}\) denote VAIOM and LightGBM NLL in bits for the same valid event. The paired gain is

\[
d_{a,t}=\ell^{\mathrm{L}}_{a,t}-\ell^{\mathrm{V}}_{a,t},
\]

so \(d_{a,t}>0\) favors VAIOM. This sign convention is the reverse of the VAIOM-minus-baseline deltas in Table~\ref{tab:main_test} and is used here so positive values denote improvement.

Because adjacent financial events are temporally dependent, individual events are not treated as independent and identically distributed (IID) bootstrap units. Events are grouped by asset and calendar month, and each asset-month is resampled as one intact block. For a period containing block set \(\mathcal{B}\), each bootstrap replicate draws \(|\mathcal{B}|\) blocks with replacement and computes the event-weighted paired mean

\[
\bar d^{*}
=
\frac{\sum_{b\in\mathcal{B}^{*}}\sum_{(a,t)\in b}d_{a,t}}
{\sum_{b\in\mathcal{B}^{*}}|b|}.
\]

The reported 95\% confidence interval is the 2.5th--97.5th percentile interval from 10,000 block-bootstrap replicates using bootstrap seed 17. Asset-month win-rate is the unweighted fraction of blocks whose within-block mean \(d_{a,t}\) is positive. Tail deltas are event-weighted paired means restricted to negative-tail buckets 1--4, positive-tail buckets 13--16, or their union. The same paired block-bootstrap protocol is applied to each of the six VAIOM seed-period evaluations. Every comparison uses the same fixed LightGBM event losses, fitted on the same pre-2024 Train sample with the same 28 single-bar fields and fixed hyperparameters. The intervals therefore measure correlated Test-sample uncertainty conditional on each fitted VAIOM and the common fixed LightGBM, while variation in model NLL across VAIOM training seeds is summarized separately by the period-specific sample standard deviation.

\subsection{Complete Capacity Rungs and Corpus Support}
\label{sec:data_capacity}

We define a \emph{complete VAIOM architecture rung} as a predefined configuration that jointly specifies decoder depth, hidden width, attention multiplicity and feed-forward capacity while retaining context length 512 and the complete return and auxiliary-head infrastructure. Table~\ref{tab:capacity_level} defines this term once for the analysis that follows. The discrete family design follows named Transformer variants and controlled model ladders \citep{devlin2019bert,dosovitskiy2021vit,bhagia2025ladders}.

\begin{table}[h]
\centering
\caption{Predefined complete VAIOM architecture rungs.}
\label{tab:capacity_level}
\begin{tabular}{lrrrrr}
\toprule
Rung & Layers & Heads & Hidden Dim & Context & Total Params \\
\midrule
0.9M & 4 & 4 & 128 & 512 & 890,086 \\
5M & 6 & 4 & 256 & 512 & 4,965,222 \\
15M & 8 & 6 & 384 & 512 & 14,726,850 \\
\bottomrule
\end{tabular}
\end{table}

\paragraph{Architecture-rung motivation.} Parameter count alone does not uniquely determine Transformer architecture. Theoretical and empirical work shows that self-attention performance depends on the interaction between depth and width, and that scaling prescriptions are sensitive to architectural shape \citep{levine2020depth,touvron2022three,alabdulmohsin2023vitshape,mcleish2025gemstones}. Compound scaling reaches the same general conclusion from a different architecture family: balancing structural dimensions can outperform scaling a single dimension \citep{tan2019efficientnet}. An intermediate parameter count therefore requires an additional depth-width-head allocation choice and is not a neutral point on a one-dimensional capacity axis.

\paragraph{Definition.} The 0.9M, 5M and 15M configurations are the evaluated complete rungs. Intermediate 2M--4M variants require a separate depth-width allocation choice and are outside this controlled comparison.

\paragraph{Corpus-support analysis.} The data-capacity analysis asks whether the fixed 1H FX corpus supports the 0.9M rung or additional model capacity. This analysis uses controlled last-position Validation runs and is not used for Test checkpoint selection.

Let \(N_r\) denote the number of non-positional parameters in complete rung \(r\) and \(U\) the number of unique valid Train windows. Specifically, \(N_r\) equals the exact total trainable parameter count minus only the learned absolute positional embedding, \(512d_r\). Input projection parameters, output and auxiliary heads, and any input-field or metadata embeddings remain included. The capacity experiment itself uses pure ContIn, so it contains no categorical metadata embeddings. The data-capacity ratio for rung \(r\) is

\[
R_{U,r}=\frac{U}{N_r}.
\]

The fixed 1H corpus contains exactly \(U=1{,}266{,}989\) unique valid Train windows. The exact denominators are

\[
\begin{aligned}
N_{0.9\mathrm{M}}&=890{,}086-512\times128=824{,}550,\\
N_{5\mathrm{M}}&=4{,}965{,}222-512\times256=4{,}834{,}150,\\
N_{15\mathrm{M}}&=14{,}726{,}850-512\times384=14{,}530{,}242.
\end{aligned}
\]

The 0.9M ratio is therefore

\[
R_{U,\min}=\frac{1{,}266{,}989}{824{,}550}=1.5366.
\]

This value is determined by the fixed corpus and the predefined 0.9M architecture. It is not selected from the Validation result.

\paragraph{Support criterion.} Corpus support for additional capacity is assessed incrementally. Under the same corpus, context, objective, auxiliary setup, optimizer budget and Validation protocol, a larger complete rung is \emph{supported} only when it improves Validation return likelihood relative to the immediately smaller complete rung. When a larger rung receives fewer unique windows per non-positional parameter and fails to produce an incremental likelihood gain, it is described as \emph{underfed} under this protocol. This is an operational classification local to the current corpus and model family.

Under this definition, \(R_U=1.5366\) is the minimum supported complete-rung ratio under the evaluated 1H protocol. It is not presented as a continuous universal threshold below which every possible architecture must fail. The Daily low-\(R_U\) experiments are used as complementary stress evidence for repeated-exposure failure and are not used to estimate a separate numerical threshold.

The complete-rung capacity analysis is conducted on the pure ContIn family. HybridContIn is a subsequent metadata-conditioning extension of the selected 0.9M backbone rung and is not used to redefine the complete-rung data-capacity ratio.

\section{Results}

This section reports the empirical results for VAIOM under the 1H FX return-distribution modeling task. The evaluation target throughout this section is the next-period normalized return bucket. All model comparisons are reported as NLL in bits per return event. Lower values are better. NLL values and deltas are rounded independently from the raw outputs to four decimal places.

The paper uses three split names consistently. \textbf{Train} is used for fitting tokenizers, model parameters and train-fitted baselines. \textbf{Validation} is used for checkpoint selection, ablation comparison and model-design diagnosis. \textbf{Test} is the fixed interval \(2025\text{-}01\text{-}01 \leq Datetime < 2026\text{-}01\text{-}01\), reported as 2025H1 and 2025H2, and is reserved for final held-out reporting of the selected main model. All ablations in Sections~\ref{sec:rq2}--\ref{sec:rq4} are evaluated on Validation only; Test is one final evaluation stage after the main model is fully selected on Validation. No Test result entered any design decision, and observations from 2026 onward are not included in the reported evaluation.

The results are organized around four research questions. First, we test whether VAIOM improves probabilistic next-return prediction beyond statistical and tree-based baselines. Second, we test whether continuous financial event input improves return-distribution modeling relative to discrete tokenized input. Third, we test whether full-sequence autoregressive supervision improves financial sequence modeling relative to last-position supervision. Fourth, we evaluate whether architecture components, including auxiliary objectives and MoMS output heads, improve the main return likelihood.

A final subsection reports data-capacity evidence as a supporting result rather than a standalone research question. It compares the predefined 0.9M, 5M and 15M rungs under the same corpus and supervision setting.

\subsection{RQ1: VAIOM Improves Test Return Likelihood beyond Statistical and Tree-Based Baselines}

RQ1 asks whether VAIOM improves held-out probabilistic next-return prediction beyond train-fitted baselines. This question is necessary before representation and architecture ablations become meaningful. If the model cannot beat basic probabilistic baselines, then later comparisons between input types, loss scopes or output heads would only compare weak variants of the same failed setup.

The selected main model is the 0.9M Hybrid Continuous Input (HybridContIn) VAIOM with a MoMS return head and Gap/VolReg/Ordinal auxiliary supervision. This configuration was selected after the pure ContIn representation study and the subsequent HybridContIn Validation comparison described in Section~\ref{sec:metadata_augmentation}. The checkpoint is selected using the Validation split and then evaluated on the 2025 Test split. Table~\ref{tab:main_test} reports results separately for 2025H1 and 2025H2. The model improves over Frequency, Markov and an identity-augmented single-bar LightGBM on both Test halves.

\begin{table}[h]
\centering
\caption{Main 2025 Test result and paired VAIOM--LightGBM uncertainty.}
\label{tab:main_test}
\resizebox{\textwidth}{!}{%
\begin{tabular}{lccccccccc}
\toprule
Model / Baseline & Input Type & Head & Aux Setup & Params & Steps & Test Bits/Event & $\Delta$ vs Frequency & $\Delta$ vs Markov & $\Delta$ vs LightGBM \\
\midrule
\multicolumn{10}{l}{\textbf{2025H1}} \\
\midrule
Frequency baseline & --- & --- & --- & --- & --- & 3.2802 & --- & --- & --- \\
Markov baseline & --- & --- & --- & --- & --- & 3.2755 & --- & --- & --- \\
LightGBM baseline & --- & --- & --- & --- & --- & 3.2057 & --- & --- & --- \\
VAIOM main model & HybridContIn & MoMS & Gap/VolReg/Ordinal & 0.9M & 5.5K & 3.1764 & -0.1037 & -0.0991 & -0.0292 \\
\midrule
\multicolumn{10}{l}{\textbf{2025H2}} \\
\midrule
Frequency baseline & --- & --- & --- & --- & --- & 3.2793 & --- & --- & --- \\
Markov baseline & --- & --- & --- & --- & --- & 3.2683 & --- & --- & --- \\
LightGBM baseline & --- & --- & --- & --- & --- & 3.1785 & --- & --- & --- \\
VAIOM main model & HybridContIn & MoMS & Gap/VolReg/Ordinal & 0.9M & 5.5K & 3.1359 & -0.1434 & -0.1324 & -0.0426 \\
\bottomrule
\end{tabular}%
}
\vspace{0.6em}

\resizebox{\textwidth}{!}{%
\begin{tabular}{lrrrrrrrrr}
\toprule
Period & Events & Blocks & Mean Gain & 95\% Interval & Block Wins & Win-Rate & Negative Tail & Positive Tail & Combined Tail \\
\midrule
2025H1 & 33,068 & 66 & +0.0292 & [+0.0205, +0.0406] & 56/66 & 84.8485\% & +0.2545 & +0.1940 & +0.2251 \\
2025H2 & 33,816 & 66 & +0.0426 & [+0.0291, +0.0626] & 61/66 & 92.4242\% & +0.3509 & +0.2428 & +0.3001 \\
\midrule
\multicolumn{10}{l}{\footnotesize Mean Gain and tail columns report \(\mathrm{NLL}_{\mathrm{LightGBM}}-\mathrm{NLL}_{\mathrm{VAIOM}}\) in bits/event; positive favors VAIOM.} \\
\bottomrule
\end{tabular}%
}
\end{table}

These Test results establish that VAIOM is not merely reproducing the unconditional return-bucket distribution, a first-order transition table or a tree-based feature model. The model compresses the held-out return distribution beyond all three baselines. This provides the empirical foundation for the remaining ablations: the later sections explain which design choices make the successful model work.

The paired asset-month analysis in the second panel of Table~\ref{tab:main_test} tests whether the LightGBM improvement is stable across the correlated Test sample. VAIOM gains 0.0292 bits/event in 2025H1 and 0.0426 bits/event in 2025H2. The corresponding 95\% block-bootstrap intervals are [0.0205, 0.0406] and [0.0291, 0.0626], both excluding zero. VAIOM wins in 56 of 66 asset-month blocks in H1 and 61 of 66 in H2. The average advantage is therefore not generated by a small number of favorable assets or months.

The gain is concentrated in difficult tail events. In 2025H1, VAIOM improves negative-tail, positive-tail and combined-tail NLL by 0.2545, 0.1940 and 0.2251 bits/event, compared with 0.0292 bits/event over all events. In 2025H2, the corresponding gains are 0.3509, 0.2428 and 0.3001 bits/event, compared with 0.0426 overall. This decomposition shows that VAIOM's aggregate compression gain includes materially better probability assignment to realized extreme-return buckets rather than only small improvements around the center of the return distribution.

A caveat on baseline fairness: VAIOM conditions on 512 bars of history, whereas LightGBM conditions on the current bar only (see Table~\ref{tab:baseline_spec}). The Test improvement over LightGBM therefore reflects the combined value of autoregressive context and learned sequence representation, not the value of the Transformer architecture in isolation. A lag-stacked or context-aggregated LightGBM would be a stronger baseline and is left for future work.

\subsubsection{Multi-Seed Robustness of the Main Test Result}
\label{sec:multiseed}

The selected configuration's Test advantage is stable across independent training runs. Seeds 17, 29 and 43 use the same 0.9M HybridContIn MoMS Gap/VolReg/Ordinal FullSeq configuration and fixed 5.5K-step budget. Every run achieves lower Test NLL than the same fixed identity-augmented single-bar LightGBM in both 2025H1 and 2025H2. All six paired asset-month block-bootstrap confidence intervals exclude zero. The individual results are reported without selecting the best seed.

\begin{table}[h]
\centering
\caption{Multi-seed robustness on the 2025 Test periods. Mean gain is \(\mathrm{NLL}_{\mathrm{LightGBM}}-\mathrm{NLL}_{\mathrm{VAIOM}}\); positive favors VAIOM.}
\label{tab:multiseed}
\resizebox{\textwidth}{!}{%
\begin{tabular}{lrrrrrr}
\toprule
Period & Seed & Model Bits/Event & LightGBM Bits/Event & Mean Gain & 95\% Interval & Block Win-Rate \\
\midrule
2025H1 & 17 & 3.1764 & 3.2057 & +0.0292 & [+0.0205, +0.0406] & 84.8485\% \\
2025H1 & 29 & 3.1880 & 3.2057 & +0.0177 & [+0.0088, +0.0284] & 71.2121\% \\
2025H1 & 43 & 3.1817 & 3.2057 & +0.0239 & [+0.0157, +0.0341] & 81.8182\% \\
\midrule
2025H2 & 17 & 3.1359 & 3.1785 & +0.0426 & [+0.0291, +0.0626] & 92.4242\% \\
2025H2 & 29 & 3.1478 & 3.1785 & +0.0307 & [+0.0183, +0.0490] & 83.3333\% \\
2025H2 & 43 & 3.1396 & 3.1785 & +0.0388 & [+0.0273, +0.0558] & 92.4242\% \\
\bottomrule
\end{tabular}
}
\end{table}

Across seeds, model NLL is $3.1820\pm0.0058$ bits/event in 2025H1 and $3.1411\pm0.0061$ in 2025H2, where dispersion is the sample standard deviation across the three training seeds. Mean gains over the common LightGBM are 0.0236 and 0.0374 bits/event, respectively. The weakest individual advantage is still positive at 0.0177 bits/event, and its confidence interval also excludes zero. The baseline win is therefore not attributable to one favorable initialization or training trajectory.

All six comparisons use exactly the same fixed LightGBM event losses. Every independently trained VAIOM therefore significantly outperforms one common baseline in both Test periods.

\subsection{RQ2: Continuous Input Improves Discrete Return Likelihood}
\label{sec:rq2}

RQ2 asks whether the pure continuous financial-event schema improves next-return distribution modeling relative to the discrete-token schema. Both variants predict the same next-period normalized return bucket and use the same bits-per-event metric, so the output objective is controlled. The input fields in this ablation are not strictly matched: DiscIn contains categorical asset, asset-class and timeframe identifiers, whereas pure ContIn contains no identity metadata or separate asset embedding.

The results show that the complete ContIn schema improves Validation return likelihood despite omitting the identity metadata available to DiscIn. This outcome is consistent with a benefit from preserving local numerical geometry, but the schema mismatch prevents attributing the full difference to continuity alone. The output remains a discrete probability distribution over return buckets, so both schemas remain directly comparable under the same likelihood objective.

\begin{table}[h]
\centering
\caption{Continuous input versus discrete input.}
\label{tab:discin_contin}
\resizebox{\textwidth}{!}{%
\begin{tabular}{lcccccl}
\toprule
Timeframe & Input Type & Params & Steps & Validation Bits/Event & $\Delta$ vs DiscIn & Result \\
\midrule
1D & DiscIn & 0.9M & 10K & 3.1743 & --- & Reference \\
1D & Pure ContIn & 0.9M & 10K & 3.1649 & -0.0094 & Win \\
1H & DiscIn & 0.9M & 50K & 3.1633 & --- & Reference \\
1H & Pure ContIn & 0.9M & 50K & 3.1536 & -0.0097 & Win \\
\bottomrule
\end{tabular}
}
\end{table}

These comparisons establish that the evaluated ContIn schema outperforms the evaluated DiscIn schema for this return-distribution task without changing the output objective. They do not constitute a field-matched causal estimate of continuity alone. VAIOM retains the discrete return-bucket likelihood while changing the representation supplied to the decoder-only backbone.

This is the key distinction between VAIOM and a conventional token-only market language model. The model remains probabilistic and token-like at the output, but it does not force all input information through categorical tokenization before representation learning.

\subsubsection{Metadata Augmentation of the Continuous-Input Model}
\label{sec:metadata_augmentation}

The primary representation experiment first established that the pure ContIn branch improves Validation return likelihood relative to DiscIn. We subsequently tested whether static market identity information remained useful after the dynamic market variables had already been represented continuously.

The HybridContIn extension retains the complete continuous event vector and adds categorical embeddings for asset identity, asset class and timeframe. Under the selected 0.9M MoMS Gap/VolReg/Ordinal FullSeq configuration, this metadata augmentation further improves Validation return likelihood from 3.1488 to 3.1465 bits/event. HybridContIn was therefore selected as the final main model and used for the reported Test evaluation.

\begin{table}[h]
\centering
\caption{Metadata augmentation of the continuous-input model.}
\label{tab:contin_meta}
\begin{tabular}{lllr}
\toprule
Variant & Dynamic Input & Identity Metadata & Validation Bits/Event \\
\midrule
Pure ContIn & 25 continuous fields & None & 3.1488 \\
HybridContIn & Same 25 fields & Asset + class + timeframe & 3.1465 \\
\bottomrule
\end{tabular}
\end{table}

This additional result does not replace the original DiscIn-versus-ContIn representation finding. Pure ContIn establishes the value of preserving the numerical geometry of dynamic market variables. The metadata extension shows that cross-asset normalization does not make static identity information redundant and that categorical market metadata can provide complementary conditioning information on top of the continuous representation.

\subsection{RQ3: Full-Sequence Supervision Improves Return Likelihood}
\label{sec:rq3}

RQ3 asks whether full-sequence autoregressive supervision improves financial return-distribution modeling relative to last-position supervision. This is a supervision-density question rather than a change in model head or input representation.

In the last-position setting, each sampled window contributes return loss only at the final position. In the full-sequence setting, every valid position in the sampled sequence contributes next-return loss under the same causal masking rule. This makes training closer to standard autoregressive pretraining: the model learns from many next-step prediction targets inside the same historical window.

Full-sequence supervision is conceptually standard in language modeling, but it is not automatically guaranteed to improve financial sequence modeling. Financial windows are highly overlapping, adjacent positions are correlated and early positions may have less effective context than the final position. Therefore, the relevant empirical question is whether denser supervision improves Validation return likelihood under the 1H FX financial corpus.

The results support FullSeq. The best FullSeq checkpoint improves over both the 50K and 300K last-position continuous-input references. However, the FullSeq sweep also shows that the optimal checkpoint occurs early; later FullSeq checkpoints can degrade. Therefore, the supported claim is not that more FullSeq steps are always better. The supported claim is that dense autoregressive supervision produces a better best Validation checkpoint when checkpoint selection is properly recalibrated. All rows in Table~\ref{tab:supervision} use the 0.9M pure-ContIn model with a MoMS head and Gap/VolReg/Ordinal auxiliary supervision.

\begin{table}[h]
\centering
\caption{Last-position versus full-sequence supervision.}
\label{tab:supervision}
\begin{tabular}{lrrr}
\toprule
Loss Scope & Steps & Validation Bits/Event & $\Delta$ vs Last-Position 50K \\
\midrule
Last-position & 50K & 3.1536 & --- \\
Last-position & 300K & 3.1542 & +0.0006 \\
Full-sequence & 5K & 3.1515 & -0.0021 \\
Full-sequence & 5.5K & 3.1488 & -0.0048 \\
\bottomrule
\end{tabular}
\par\vspace{3pt}
{\footnotesize All runs use the 0.9M model. Gradient accumulation is four except for the 300K last-position reference.}
\end{table}

The 5K and 5.5K FullSeq checkpoints show that full-sequence supervision improves the best Validation likelihood relative to last-position supervision. The 5.5K checkpoint is the strongest observed Validation model in this comparison. This supports the use of dense autoregressive supervision for financial return-distribution modeling.

Because each sampled window supplies many correlated supervised positions under FullSeq, the model consumes useful training signal faster than in the last-position setting. The Validation optimum therefore shifts earlier, and FullSeq should not inherit the 50K / 300K step schedule from last-position training.

\paragraph{Optimizer steps versus supervised target exposure.} Optimizer steps are not directly comparable across the two supervision scopes, because each step under FullSeq supervises many more target positions than under last-position supervision. For this target-exposure calculation, the effective batch is \(B_{\mathrm{eff}}=128\) (batch 32, grad-accum 4) and context length is \(C=512\):

Under last-position supervision, \(Q=1\), so the 50K reference run exposes
\[
D_{\mathrm{raw}}^{\mathrm{last}}=S\,B_{\mathrm{eff}}\,Q=50000\times128\times1=6.4\text{M}
\]
supervised targets. Under FullSeq, each 512-length window contains approximately \(Q\approx511\) valid target positions in the 1H FX corpus (masks are rare in clean FX and spot-metal bars), so the 5.5K run exposes
\[
D_{\mathrm{raw}}^{\mathrm{full}}=5500\times128\times511\approx360\text{M}
\]
supervised targets. The 5.5K FullSeq checkpoint therefore sees roughly \(56\times\) more supervised targets than the 50K last-position reference, despite using \(9\times\) fewer optimizer steps. The ``fewer steps'' observation concerns optimizer-step and wall-clock efficiency, not supervised-target budget. FullSeq converts target exposure into Validation improvement more effectively on a per-optimizer-step basis, but the experiment does not show that it requires fewer total supervised targets. Adjacent positions are highly correlated, so the \(56\times\) raw target count overstates the effective information gain; this is consistent with the early Validation optimum and the degradation of later FullSeq checkpoints.

\subsection{RQ4: Architecture Components Improve the Main Return Objective}
\label{sec:rq4}

RQ4 asks whether VAIOM architecture components improve the main return likelihood. This section evaluates two component groups: auxiliary objectives and the MoMS output head. Both are judged only by their effect on return-bucket likelihood. Auxiliary losses are not treated as separate success metrics, and MoMS latent states are not assigned economic labels.

\subsubsection{Auxiliary Heads Provide Component-Dependent Return-NLL Improvements}

VAIOM uses auxiliary objectives as representation-shaping signals. The auxiliary system contains Gap, volatility-regime and ordinal supervision. These heads are not evaluated by their own standalone accuracy in the main results. Their value is measured by whether they improve the main return likelihood.

The available component-level auxiliary ablation is the 1H DiscIn sweep. This is reported as auxiliary-head evidence, not as a ContIn-only causal claim. The results support auxiliary supervision as a representation-shaping system, but the contribution is component-dependent. Gap auxiliary supervision provides a mild positive improvement. Volatility-regime supervision does not improve return NLL when isolated as a standalone auxiliary head. Ordinal supervision provides the clearest positive contribution when added to the Gap/VolReg auxiliary reference.

\begin{table}[h]
\centering
\caption{Component-level auxiliary-head ablations.}
\label{tab:aux_ablation}
\small
\textit{Panel A: 0.9M isolated auxiliary objectives}\\[3pt]
\begin{tabular}{lccccrr}
\toprule
Input & Gap & VolReg & Ordinal & Weight & Steps & Validation Bits/Event \\
\midrule
DiscIn & No & No & No & 0.00 & 45K & 3.1658 \\
DiscIn & Yes & No & No & 0.10 & 45K & 3.1648 \\
DiscIn & No & Yes & No & 0.10 & 45K & 3.1659 \\
\bottomrule
\end{tabular}

\vspace{6pt}
\textit{Panel B: 5M Ordinal increment over Gap/VolReg}\\[3pt]
\begin{tabular}{lccccrrr}
\toprule
Input & Gap & VolReg & Ordinal & Weight & Steps & Validation Bits/Event & $\Delta$ \\
\midrule
DiscIn & Yes & Yes & No & 0.10 & 180K & 3.1719 & --- \\
DiscIn & Yes & Yes & Yes & 0.10 & 180K & 3.1673 & -0.0046 \\
\bottomrule
\end{tabular}
\end{table}

The empirical claim is therefore not that every auxiliary target independently improves return prediction. The supported claim is that auxiliary supervision improves the main return objective as a representation-shaping system, with positive evidence from Gap and Ordinal supervision and neutral-to-negative isolated evidence from Volatility-regime supervision.

The ordinal result is especially important because return buckets are ordered. A model that assigns probability mass to a nearby bucket should not be treated the same as one that assigns probability mass to a distant tail bucket. The ordinal auxiliary head introduces this geometry during training while preserving the main categorical likelihood for evaluation.

\subsubsection{MoMS Improves over Independent Categorical Heads}

The second architecture component is the MoMS return head. MoMS is tested against an independent categorical return head to determine whether a mixture-parameterized output distribution improves return likelihood.

The results show that MoMS provides positive marginal value in matched head ablations. This supports the use of a mixture-style output distribution for return-bucket prediction. However, the result should not be interpreted as proof that the learned mixture components correspond to economically identifiable market states. The empirical claim is narrower: mixture-parameterized return heads improve Validation return likelihood relative to independent categorical heads in matched comparisons.

\begin{table}[h]
\centering
\caption{MoMS versus independent categorical head.}
\label{tab:moms_vs_indep}
\resizebox{\textwidth}{!}{%
\begin{tabular}{lccccccl}
\toprule
Timeframe & Input Type & Head & Params & Steps & Validation Bits/Event & $\Delta$ vs Independent & Result \\
\midrule
1H & Continuous & Independent & 0.9M & 10K & 3.1651 & --- & Reference \\
1H & Continuous & MoMS & 0.9M & 10K & 3.1563 & -0.0088 & MoMS improves \\
4H & Discrete & Independent & 5M & 30K & 3.1887 & --- & Reference \\
4H & Discrete & MoMS & 5M & 30K & 3.1834 & -0.0053 & MoMS improves \\
\bottomrule
\end{tabular}%
}
\end{table}

MoMS is therefore retained as the output-head design in the main VAIOM configuration. Its value is architectural and probabilistic: it allows the model to express the next-return distribution as a mixture of latent predictive components. The paper does not rely on naming or interpreting these components as explicit market regimes.

\subsection{Supporting Capacity Analysis}

Table~\ref{tab:data_capacity} reports the controlled complete-rung comparison. The three 1H runs use the same continuous corpus, 512-bar context, MoMS return head, Gap/VolReg/Ordinal objectives and 300K last-position optimizer steps. The changed variable is the complete capacity rung.

This capacity comparison is conducted on the pure ContIn family. HybridContIn is a later metadata-conditioning extension of the selected 0.9M backbone rung and does not redefine \(N\), \(R_U\) or the complete-rung family used here.

The 0.9M rung achieves the strongest Validation likelihood at 3.1542 bits/event. Increasing capacity to 5M reduces \(R_U\) from 1.5366 to 0.2621 and worsens Validation likelihood by 0.0038 bits/event. Increasing capacity to 15M reduces \(R_U\) to 0.0872 and worsens Validation likelihood by 0.0065 bits/event relative to 0.9M.

Under the incremental criterion in Section~\ref{sec:data_capacity}, added parameter capacity is not converted into better Validation compression. The result is local to the current corpus, architecture family, objective and last-position diagnostic. Appendix~\ref{app:frequency_capacity} reports the Daily stress evidence and frequency-selection development path.

\begin{table}[h]
\centering
\caption{Controlled 1H complete-rung corpus-support comparison.}
\label{tab:data_capacity}
\begin{tabular}{lrrrrr}
\toprule
Rung & Non-positional $N$ & $R_U$ & Steps & Validation NLL & $\Delta$ vs 0.9M \\
\midrule
0.9M & 824,550 & 1.5366 & 300K & 3.1542 & --- \\
5M & 4,834,150 & 0.2621 & 300K & 3.1580 & +0.0038 \\
15M & 14,530,242 & 0.0872 & 300K & 3.1607 & +0.0065 \\
\bottomrule
\end{tabular}
\end{table}

\subsection{Summary of Empirical Findings}

The selected HybridContIn model improves 2025 Test likelihood beyond all three baselines across three training seeds. Validation comparisons favor the evaluated continuous-input branch, FullSeq supervision and HybridContIn metadata augmentation. Component ablations provide positive but component-dependent evidence for auxiliary supervision and favor MoMS over an independent categorical head. The capacity comparison is supporting evidence: among the evaluated 1H rungs, 0.9M achieves the strongest Validation likelihood.

\section{Discussion}

\subsection{Continuous Input and Categorical Output}

VAIOM represents dynamic financial events as continuous vectors while retaining a categorical return-distribution target. The two sides of the task need not share one representation. Continuous input preserves numerical relationships among returns, gaps, volatility and calendar variables; categorical output retains cross-entropy training, bits-per-event evaluation and direct comparison with probabilistic baselines.

The ContIn--DiscIn result is an end-to-end schema comparison rather than a field-matched estimate of continuity alone. DiscIn contains identity metadata that pure ContIn omits, so the observed difference cannot be assigned uniquely to bucketization. Nevertheless, pure ContIn performs better despite this omission. HybridContIn then improves on pure ContIn by adding asset, asset-class and timeframe embeddings. Volatility normalization therefore creates useful cross-asset comparability without making static identity information redundant.

This formulation differs from a fully tokenized market language model. Dynamic market variables retain their continuous geometry, while categorical metadata provides lightweight conditioning and the return target remains probabilistic. The evidence does not compare categorical output with a continuous-output density model; that question remains open.

\subsection{Full-Sequence Supervision Changes the Training Regime}

FullSeq uses standard causal autoregressive supervision \citep{vaswani2017attention}. Decoder-only time-series models use the same principle \citep{das2024timesfm,rasul2023lagllama}. Its effect here is empirical: supervising every valid causal position improves the best Validation likelihood relative to last-position training on the 1H corpus.

The result concerns supervision density, not model architecture or total target budget. A FullSeq window contributes hundreds of correlated targets, so its Validation optimum occurs much earlier in optimizer-step terms even though it receives more raw target exposures. FullSeq therefore requires its own checkpoint schedule. Later checkpoints can degrade as repeated, highly dependent positions add less effective information than their nominal count suggests.

\subsection{Auxiliary Objectives as Representation Shaping}

Auxiliary objectives are useful only when they improve the main return likelihood. Gap and Ordinal supervision provide positive evidence, whereas isolated VolReg supervision does not improve return NLL under the fixed-weight schedule. The result therefore supports auxiliary representation shaping as a component-dependent system, not a claim that every auxiliary target is independently beneficial.

Ordinal supervision is well matched to the ordered return buckets: it exposes distance structure during training while leaving the evaluation objective categorical. VolReg is more persistent and easier to predict than next return. Its useful gradient signal may saturate earlier than the main task, after which a constant weight can become redundant or conflicting.

\subsection{MoMS as an Output-Head Hypothesis}

MoMS improves Validation likelihood over independent categorical heads in the matched comparisons. This supports mixture-parameterized categorical output as an architectural choice. It does not identify economic regimes. The latent components are permutation-invariant and may split, merge or change role across seeds and checkpoints; interpreting them requires diagnostics beyond likelihood improvement.

\subsection{Small-Model Strength and Corpus Support}

The controlled 1H comparison favors the 0.9M rung: neither 5M nor 15M converts additional capacity into better Validation compression. Under the operational criterion used here, the larger rungs are underfed by the present corpus and training protocol. This is a local corpus-support result, not a universal parameter threshold.

The Daily--4H--1H development path in Appendix~\ref{app:frequency_capacity} provides supporting context. Denser corpora produced more stable Validation behavior, but nominal bar count is not equivalent to independent financial information because adjacent windows overlap and finer bars can add microstructure noise. Broader asset coverage, longer history and genuinely incremental observations are alternative ways to increase effective corpus support.

\subsection{Limitations}

The evidence is restricted to one-hour FX and spot-metal data from one source. It does not establish generality across equities, futures, cryptoassets, options, order flow or mixed-frequency corpora. The bucketized target also leaves open whether continuous density, quantile or hybrid output heads would retain more information, particularly in the tails.

The representation ablation is not field-matched, and the LightGBM comparison is intentionally limited to single-bar features while VAIOM receives 512 bars of context. The baseline result therefore measures the combined value of sequence context and learned representation, not Transformer architecture in isolation. Lag-stacked tree models and finance-specific sequence baselines remain necessary comparisons.

Multi-seed and block-bootstrap analyses strengthen the final baseline comparison, but component ablations remain single-seed controlled experiments. Asset-month resampling preserves dependence within each block but cannot remove all cross-asset or longer-horizon dependence. MoMS components are not economically identified, and the capacity conclusion remains specific to the evaluated family.

Finally, likelihood compression is not trading profitability. Deployment would require mapping predicted distributions to positions, restoring raw-return scale and accounting for transaction costs, liquidity, execution, turnover, portfolio construction and risk management.

\subsection{Future Work}

Future evaluation should add lag-stacked and sequence-aware baselines, continuous or hybrid output distributions, broader markets and longer histories. Auxiliary objectives should receive task-specific schedules: VolReg-specific early stopping, head freezing or loss-weight decay can test whether its weak isolated contribution reflects optimization timing rather than an uninformative target. MoMS analysis should examine component usage, separation and stability across seeds. These extensions would test which gains survive stronger baselines, richer corpora and less restrictive output assumptions.

\section{Conclusion}

VAIOM adapts decoder-only sequence modeling to continuous financial observations by separating input representation from output likelihood. Continuous multivariate market-event vectors preserve numerical structure, while a categorical distribution over the next volatility-normalized return bucket supports causal cross-entropy training and held-out probabilistic evaluation.

The selected 0.9M HybridContIn model consistently outperforms Frequency, Markov and the fixed identity-augmented single-bar LightGBM on both 2025 Test halves. Across three seeds, every VAIOM run beats LightGBM; the canonical paired gains are 0.0292 and 0.0426 bits per event. Validation experiments favor the evaluated continuous-input branch, HybridContIn metadata augmentation, FullSeq supervision and MoMS, while auxiliary-objective gains are component-dependent. The supporting capacity comparison favors the 0.9M rung over the evaluated 5M and 15M alternatives on the present corpus.

These findings concern likelihood compression in one 1H FX setting, not universal architecture scaling or trading profitability. Stronger sequence baselines, field-matched representation studies, alternative output distributions and broader corpora are required to determine how far the formulation generalizes.

\clearpage
\appendix
\section{Full Experimental Configuration}
\label{app:full_configuration}

This appendix collects the evidence-bearing runs used in the paper. It excludes smoke tests, failed jobs, infrastructure retries and path-repair runs because they do not provide scientific evidence. Unless stated otherwise, likelihoods are Validation negative log-likelihood (NLL) in bits per event on 2024H2. Test results are reported separately in Appendix~\ref{app:period_results}.

\subsection{Selected Main Configuration}

\begin{table}[h]
\centering
\caption{Exact configuration of the selected Hybrid Continuous Input (HybridContIn) model.}
\label{tab:app_main_config}
\begin{tabular}{p{0.27\textwidth}p{0.66\textwidth}}
\toprule
Component & Setting \\
\midrule
Corpus / bar frequency & V0.2 FX and spot-metal corpus / 1H \\
Dynamic input & 25 continuous financial-event fields \\
Static metadata & Asset, asset class and timeframe; 8-d embedding each \\
Fusion & 25-to-128 MLP; concatenate 24 metadata dimensions; 152-to-128 MLP \\
Backbone & 4 pre-norm decoder blocks, 4 heads, width 128, FFN width 512 \\
Position / context & Learned absolute position embedding / 512 bars \\
Return head & Mixture of Market States (MoMS), 4 latent components, 16 buckets \\
Auxiliary objectives & Gap / VolReg / Ordinal; weights 0.1000 / 0.1000 / 0.1000 \\
Supervision & Full-sequence (FullSeq) valid-position supervision \\
Parameters / steps & 0.9116M total parameters / 5.5K optimizer steps \\
Optimization & AdamW, LR 0.0003, weight decay 0.0100, constant schedule \\
Batching & Micro-batch 32, gradient accumulation 4, effective batch 128 \\
Regularization & Dropout 0.1000, gradient-norm clipping 1.0000 \\
Numerics & FP16 automatic mixed precision and data parallelism \\
Training seeds & 17 (canonical), 29 and 43 (robustness) \\
Selection rule & Lowest 2024H2 Validation return NLL in the prespecified sweep \\
\bottomrule
\end{tabular}
\end{table}

\subsection{Evidence-Bearing Validation Runs}

Table~\ref{tab:app_validation_matrix} indexes the formal comparisons used in the main argument. ``Aux'' abbreviates the three named auxiliary targets only; VolReg is never raw volume. Rows from different panels should not be treated as matched unless the corresponding Results section identifies them as a controlled comparison.

\begin{table}[h]
\centering
\caption{Formal Validation runs used in the reported comparisons.}
\label{tab:app_validation_matrix}
\resizebox{\textwidth}{!}{%
\begin{tabular}{lllclrrl}
\toprule
Comparison & Timeframe & Input / head & Aux / scope & Params & Steps & Validation NLL & Role \\
\midrule
Representation & 1D & DiscIn / MoMS & None / last & 0.9M & 10K & 3.1743 & Reference \\
Representation & 1D & Pure ContIn / MoMS & None / last & 0.9M & 10K & 3.1649 & Comparator \\
Representation & 1H & DiscIn / MoMS & Gap/VolReg/Ord / last & 0.9M & 50K & 3.1633 & Reference \\
Representation & 1H & Pure ContIn / MoMS & Gap/VolReg/Ord / last & 0.9M & 50K & 3.1536 & Comparator \\
\midrule
Metadata & 1H & Pure ContIn / MoMS & Gap/VolReg/Ord / FullSeq & 0.9M & 5.5K & 3.1488 & Reference \\
Metadata & 1H & HybridContIn / MoMS & Gap/VolReg/Ord / FullSeq & 0.9M & 5.5K & 3.1465 & Selected, seed 17 \\
Metadata & 1H & HybridContIn / MoMS & Gap/VolReg/Ord / FullSeq & 0.9M & 5.5K & 3.1525 & Robustness, seed 29 \\
Metadata & 1H & HybridContIn / MoMS & Gap/VolReg/Ord / FullSeq & 0.9M & 5.5K & 3.1450 & Robustness, seed 43 \\
\midrule
Auxiliary & 1H & DiscIn / MoMS & None / last & 0.9M & 45K & 3.1658 & Reference \\
Auxiliary & 1H & DiscIn / MoMS & Gap / last & 0.9M & 45K & 3.1648 & Isolated Gap \\
Auxiliary & 1H & DiscIn / MoMS & VolReg / last & 0.9M & 45K & 3.1659 & Isolated VolReg \\
Auxiliary & 1H & DiscIn / MoMS & Gap/VolReg / last & 5M & 180K & 3.1719 & Ordinal reference \\
Auxiliary & 1H & DiscIn / MoMS & Gap/VolReg/Ord / last & 5M & 180K & 3.1673 & Ordinal 0.1000 \\
\midrule
Return head & 1H & Pure ContIn / Independent & Gap/VolReg/Ord / FullSeq & 0.9M & 10K & 3.1651 & Reference \\
Return head & 1H & Pure ContIn / MoMS & Gap/VolReg/Ord / FullSeq & 0.9M & 10K & 3.1563 & Comparator \\
Return head & 4H & DiscIn / Independent & Gap/VolReg / last & 5M & 30K & 3.1887 & Reference \\
Return head & 4H & DiscIn / MoMS & Gap/VolReg / last & 5M & 30K & 3.1834 & Comparator \\
\midrule
Capacity & 1H & Pure ContIn / MoMS & Gap/VolReg/Ord / last & 0.9M & 300K & 3.1542 & Rung \\
Capacity & 1H & Pure ContIn / MoMS & Gap/VolReg/Ord / last & 5M & 300K & 3.1580 & Rung \\
Capacity & 1H & Pure ContIn / MoMS & Gap/VolReg/Ord / last & 15M & 300K & 3.1607 & Rung \\
\bottomrule
\end{tabular}%
}
\end{table}

\FloatBarrier

\section{Corpus and Tokenizer Audit}
\label{app:corpus_audit}

\subsection{Chronological Splits and Recorded Sample Counts}

\begin{table}[h]
\centering
\caption{Recorded sample counts and split roles for the main 1H study.}
\label{tab:app_split_counts}
\resizebox{\textwidth}{!}{%
\begin{tabular}{llllr}
\toprule
Interval & Boundary & Fitting permitted & Role & Valid events/windows \\
\midrule
Train & \(t<2024\text{-}01\text{-}01\) & Yes & Parameter estimation & 1,266,989 \\
Buffer & 2024H1 & No & Causal recursive history only & --- \\
Validation & 2024H2 & No & Selection and ablation & 33,911 \\
Test & 2025H1 & No & Held-out reporting & 33,068 \\
Test & 2025H2 & No & Held-out reporting & 33,816 \\
Reserved & \(t\geq2026\text{-}01\text{-}01\) & No & Future holdout & --- \\
\bottomrule
\end{tabular}%
}
\end{table}

\FloatBarrier
Train determines bucket edges, categorical maps and serialized preprocessing rules. Exponentially weighted moving-average (EWMA) states continue causally through the buffer, Validation and Test using past-only observations, but no parameter is refit outside Train.

\subsection{Asset Coverage}

``Rows'' in Table~\ref{tab:asset_coverage} counts stored non-synthesized 1H bars across the archive. ``Valid Train windows'' counts 512-bar context endpoints before 2024 that pass the target and mask rules in Section~\ref{sec:buckets}.

\begin{table}[h]
\centering
\caption{Per-asset 1H archive coverage and valid Train-window counts.}
\label{tab:asset_coverage}
\resizebox{\textwidth}{!}{%
\begin{tabular}{llrrll}
\toprule
Asset & Class & Rows & Valid Train Windows & First Bar & Last Bar \\
\midrule
AUDUSD & FX & 140,325 & 122,905 & 2003-08-04 00:00 & 2026-03-13 20:00 \\
EURUSD & FX & 142,744 & 126,950 & 2003-05-05 00:00 & 2026-03-13 20:00 \\
GBPUSD & FX & 142,731 & 126,812 & 2003-05-05 00:00 & 2026-03-13 20:00 \\
NZDUSD & FX & 141,123 & 124,213 & 2003-08-04 00:00 & 2026-03-13 20:00 \\
USDCAD & FX & 141,162 & 124,684 & 2003-08-04 00:00 & 2026-03-13 20:00 \\
USDCNY & FX & 85,008 & 70,346 & 2012-06-27 00:00 & 2026-03-13 20:00 \\
USDHKD & FX & 118,417 & 95,594 & 2007-03-14 00:00 & 2026-03-13 20:00 \\
USDJPY & FX & 142,714 & 124,642 & 2003-05-05 00:00 & 2026-03-13 20:00 \\
USDSGD & FX & 133,072 & 113,855 & 2004-11-17 00:00 & 2026-03-13 20:00 \\
XAGUSD & Spot metal & 135,869 & 112,262 & 2003-08-08 15:00 & 2026-03-09 23:00 \\
XAUUSD & Spot metal & 138,622 & 124,726 & 2003-05-05 00:00 & 2026-03-09 23:00 \\
\midrule
Total & --- & 1,461,787 & 1,266,989 & --- & --- \\
\bottomrule
\end{tabular}%
}
\end{table}

\FloatBarrier
\subsection{Complete Input Schemas}

\begin{table}[h]
\centering
\caption{Complete DiscIn field list for the reported 1H corpus.}
\label{tab:discin_fields}
\begin{tabular}{rlll}
\toprule
Index & Field & Index & Field \\
\midrule
1 & \texttt{return\_bucket} & 9 & \texttt{day\_of\_month} \\
2 & \texttt{gap\_bucket} & 10 & \texttt{day\_of\_week} \\
3 & \texttt{vol\_regime} & 11 & \texttt{day\_of\_year} \\
4 & \texttt{asset\_id} & 12 & \texttt{hour} \\
5 & \texttt{asset\_class\_id} & 13 & \texttt{minute} \\
6 & \texttt{timeframe\_id} & 14 & \texttt{second} \\
7 & \texttt{years\_since\_2000} & 15 & \texttt{mask\_any} \\
8 & \texttt{month\_of\_year} & --- & --- \\
\bottomrule
\end{tabular}
\end{table}

\begin{table}[h]
\centering
\caption{Complete numerical field list for pure ContIn and HybridContIn.}
\label{tab:contin_fields}
\begin{tabular}{rlll}
\toprule
Index & Field & Index & Field \\
\midrule
1 & \texttt{ret\_z} & 14 & \texttt{hour\_sin} \\
2 & \texttt{gap\_z} & 15 & \texttt{hour\_cos} \\
3 & \texttt{relative\_log\_vol} & 16 & \texttt{minute\_sin} \\
4 & \texttt{sigma\_through\_t} & 17 & \texttt{minute\_cos} \\
5 & \texttt{years\_since\_2000\_norm} & 18 & \texttt{second\_sin} \\
6 & \texttt{month\_sin} & 19 & \texttt{second\_cos} \\
7 & \texttt{month\_cos} & 20 & \texttt{mask\_missing} \\
8 & \texttt{day\_of\_month\_sin} & 21 & \texttt{mask\_stale} \\
9 & \texttt{day\_of\_month\_cos} & 22 & \texttt{mask\_bad\_data} \\
10 & \texttt{day\_of\_week\_sin} & 23 & \texttt{mask\_insufficient\_history} \\
11 & \texttt{day\_of\_week\_cos} & 24 & \texttt{mask\_scale\_zero} \\
12 & \texttt{day\_of\_year\_sin} & 25 & \texttt{mask\_any} \\
13 & \texttt{day\_of\_year\_cos} & --- & --- \\
\bottomrule
\end{tabular}
\end{table}

\FloatBarrier
HybridContIn preserves all 25 numerical fields and adds the three categorical metadata identifiers through separate learned embeddings. Raw volume is disabled for the FX and spot-metal corpus and is not among these fields.

\subsection{Return Buckets and Validity Rules}
\label{app:bucket_validity}

The reported tokenizer uses 16 non-missing return buckets. Its Train-fitted edges are
\[
(-\infty,-8,-5,-3,-2,-1.0085,-0.6241,-0.2697,
0.0118,0.2833,0.6362,1.0206,2,3,5,8,+\infty).
\]
Buckets 1--4 and 13--16 define the negative and positive tails used in the Test decomposition. Index 0 is reserved for invalid or missing targets and never enters a likelihood denominator.

The Gap target uses the 20-span past-only EWMA gap scale defined in Methods, with floor \(10^{-8}\). The same signed-edge procedure is fitted on Train: fixed cutoffs \((-8,-5,-3,-2,2,3,5,8)\) are combined with the unique interior quantiles. The serialized 1H Gap edges are

\[
(-\infty,-8,-5,-3,-2,-0.6776,-0.3059,0,
0.3195,0.7012,2,3,5,8,+\infty).
\]

Duplicate Train quantiles at zero collapse during edge deduplication, so these 15 edges define 14 non-missing Gap buckets. Missing or invalid Gap targets use index 0.

\begin{table}[h]
\centering
\caption{Serialized row and target validity rules.}
\label{tab:app_mask_rules}
\resizebox{\textwidth}{!}{%
\begin{tabular}{ll}
\toprule
Flag & Exact rule \\
\midrule
Missing & Open or Close is null or non-positive \\
Stale & Previous close exists and close-to-close log return equals zero \\
Bad data & Absolute close return or open gap exceeds 1.0000 in log units \\
Insufficient history & Fewer than 20 preceding clean returns \\
Scale zero & Past-only volatility scale is at or below \(10^{-8}\) \\
Mask any & Logical OR of the five row-level flags above \\
Target bad & Non-finite, exactly zero or excessive next return, or invalid next Open/Close \\
FullSeq return base & Return target index \(>0\), current \texttt{mask\_any} false and \texttt{target\_bad} false \\
FullSeq auxiliary & Return base true and task-specific target index \(>0\) \\
\bottomrule
\end{tabular}%
}
\end{table}

\FloatBarrier
The current experiment registry records these rules and valid-window counts, but not a complete split-by-asset table of mask rates, exact-zero rates or bucket occupancy. No such quantities are reconstructed or inferred here.

\FloatBarrier
\FloatBarrier

\section{Training and Checkpoint Diagnostics}
\label{app:training_diagnostics}

The available diagnostics are checkpoint-level Validation evaluations rather than continuously sampled curves. They nevertheless show the selection trajectory and the onset of overtraining without requiring interpolation between recorded checkpoints.

\subsection{Selected FullSeq Sweep}

\begin{table}[h]
\centering
\caption{Pure-ContIn FullSeq checkpoint sweep used to select the 5.5K reference before metadata augmentation.}
\label{tab:app_fullseq_sweep}
\begin{tabular}{lrrrr}
\toprule
Input & Params & Steps & Validation NLL & \(\Delta\) vs 5.5K \\
\midrule
Pure ContIn & 0.9M & 5.0K & 3.1515 & +0.0028 \\
Pure ContIn & 0.9M & 5.5K & 3.1488 & 0.0000 \\
Pure ContIn & 0.9M & 6.0K & 3.1497 & +0.0009 \\
Pure ContIn & 0.9M & 6.5K & 3.1510 & +0.0023 \\
Pure ContIn & 0.9M & 7.0K & 3.1491 & +0.0003 \\
Pure ContIn & 0.9M & 10K & 3.1563 & +0.0075 \\
Pure ContIn & 0.9M & 20K & 3.2760 & +0.1272 \\
\bottomrule
\end{tabular}
\end{table}

\FloatBarrier
The local optimum is visible within the 500-step checkpoint spacing. Validation degradation at 10K and 20K also shows why the earlier last-position training budget was not transferred to FullSeq.

\subsection{Long-Horizon 1H Checkpoint Path}

\begin{table}[h]
\centering
\caption{Recorded 1H 5M DiscIn Gap/VolReg last-position checkpoint trajectory.}
\label{tab:app_1h_long_trajectory}
\resizebox{\textwidth}{!}{%
\begin{tabular}{lrrrrrrrrrr}
\toprule
Params & 50K & 75K & 100K & 125K & 150K & 180K & 250K & 300K & 350K & 400K \\
\midrule
5M & 3.2072 & 3.1914 & 3.1772 & 3.1741 & 3.1677 & 3.1719 & 3.1679 & 3.1631 & 3.1637 & 3.1679 \\
\bottomrule
\end{tabular}%
}
\end{table}

\FloatBarrier
This historical trajectory reaches its best recorded checkpoint at 300K and then degrades. It supports Validation-based checkpoint selection but is not directly comparable to the later 0.9M FullSeq sweep because both supervision density and architecture differ.

\subsection{Daily Repeated-Exposure Trajectory}

\begin{table}[h]
\centering
\caption{Daily 5M checkpoint trajectories from the prolonged repeated-exposure stress run.}
\label{tab:app_daily_trajectory}
\begin{tabular}{llrrrrr}
\toprule
Head & Params & 20K & 30K & 40K & 200K & Best recorded step \\
\midrule
MoMS & 5M & 3.2072 & 3.3086 & 3.8671 & 8.9647 & 20K \\
Independent & 5M & 3.2541 & 3.4837 & 4.1994 & 8.8181 & 20K \\
\bottomrule
\end{tabular}
\end{table}

\FloatBarrier
Both output heads deteriorate after 20K, despite continued optimization on Train. The common trajectory is evidence of corpus-level repeated exposure rather than a failure unique to the MoMS head.

\FloatBarrier
\FloatBarrier

\section{Additional Ablations and Robustness Checks}
\label{app:additional_ablations}

\subsection{Frequency-Selection Development Path}
\label{app:frequency_capacity}

The 1H corpus was selected through a progressive development path rather than a matched causal frequency ablation. Daily provided 1,726 Validation windows, 4H provided 8,882 and 1H provided 33,911. Table~\ref{tab:app_frequency_path} records the strongest representative checkpoint at each stage. Architecture and optimizer exposure differ, so the table documents development evidence rather than an isolated frequency effect.

\begin{table}[h]
\centering
\caption{Daily-to-4H-to-1H corpus-density development path.}
\label{tab:app_frequency_path}
\resizebox{\textwidth}{!}{%
\begin{tabular}{lrrllrr}
\toprule
Timeframe & Validation windows & Params & Input & Head / scope & Steps & Validation NLL \\
\midrule
1D & 1,726 & 0.9M & Pure ContIn & MoMS / last & 10K & 3.1649 \\
4H & 8,882 & 5M & DiscIn & MoMS / last & 30K & 3.1834 \\
1H & 33,911 & 0.9M & Pure ContIn & MoMS / FullSeq & 5.5K & 3.1488 \\
\bottomrule
\end{tabular}%
}
\end{table}

\FloatBarrier
The progression does not imply that finer bars are always better. It motivates a denser corpus with more effective unique information; finer bars can also increase dependence and microstructure noise.

\subsection{MoMS State Count, Ordinal Weight and Context Length}

\begin{table}[h]
\centering
\caption{Additional MoMS, Ordinal and context diagnostics. Each panel identifies its local reference.}
\label{tab:app_hyper_ablations}
\resizebox{\textwidth}{!}{%
\begin{tabular}{lllrrrl}
\toprule
Panel & Setting & Local comparison & Params & Steps & Validation NLL & Result \\
\midrule
MoMS states & \(K_{\mathrm{states}}=4\) & Ordinal 0.1000 reference & 5M & 180K & 3.1673 & Reference \\
MoMS states & \(K_{\mathrm{states}}=8\) & Same configuration & 5M & 180K & 3.1707 & Worse \\
MoMS states & \(K_{\mathrm{states}}=16\) & Same configuration & 5M & 180K & 3.1659 & Better by 0.0014 \\
\midrule
Ordinal weight & 0.0000 & Gap/VolReg reference & 5M & 180K & 3.1719 & Reference \\
Ordinal weight & 0.0500 & Same configuration & 5M & 180K & 3.1695 & Better by 0.0025 \\
Ordinal weight & 0.1000 & Same configuration & 5M & 180K & 3.1673 & Better by 0.0046 \\
\midrule
Context & CTX512 & Gap/VolReg, no Ordinal & 5M & 300K & 3.1631 & Local reference \\
Context & CTX1024 & Gap/VolReg, no Ordinal & 5M & 175K & 3.1604 & Best CTX1024 checkpoint \\
Context & CTX1024 & Gap/VolReg, no Ordinal & 5M & 250K & 3.1691 & Later checkpoint \\
Context & CTX1024 & Gap/VolReg, no Ordinal & 5M & 300K & 3.1663 & Later checkpoint \\
\bottomrule
\end{tabular}%
}
\end{table}

\FloatBarrier
The state-count and Ordinal panels are matched at 180K. The context panel is only a checkpoint diagnostic because the best recorded CTX512 and CTX1024 evaluations occur at different optimizer steps.

\subsection{Additional Capacity Diagnostics}

\begin{table}[h]
\centering
\caption{Daily low-\(R_U\) repeated-exposure stress evidence; \(U=0.0602\mathrm{M}\) for every row.}
\label{tab:daily_capacity_appendix}
\begin{tabular}{lrrrrl}
\toprule
Run & Params (\(N\)) & \(R_U\) & Steps & Validation NLL & Evidence \\
\midrule
0.9M ContIn NoAux & 0.820M & 0.0734 & 10K & 3.1649 & Short-run reference \\
5M MoMS Aux & 4.893M & 0.0123 & 200K & 8.9647 & Validation collapse \\
5M Independent Aux & 4.880M & 0.0123 & 200K & 8.8181 & Validation collapse \\
\bottomrule
\end{tabular}
\end{table}

\begin{table}[h]
\centering
\caption{Exploratory 10-minute capacity checkpoints on the sampled Validation evaluation.}
\label{tab:app_10min_capacity}
\begin{tabular}{lrrrl}
\toprule
Input / head & Params & Steps & Validation NLL & Use \\
\midrule
Pure ContIn / MoMS & 5M & 5K & 3.2193 & Early checkpoint \\
Pure ContIn / MoMS & 5M & 50K & 3.1922 & Capacity comparator \\
Pure ContIn / MoMS & 0.9M & 50K & 3.1889 & Capacity comparator \\
Pure ContIn / MoMS & 0.9M & 100K & 3.1829 & Later checkpoint \\
\bottomrule
\end{tabular}
\end{table}

\FloatBarrier
At the matched 50K budget, the 0.9M model is better by 0.0033 bits/event. These runs are exploratory and do not redefine the main 1H complete-rung analysis.

\FloatBarrier
\FloatBarrier

\section{Per-Period and Block-Level Results}
\label{app:period_results}

No separate per-asset NLL table is available in the formal registry. The retained decomposition is by Test half and by complete asset-month blocks, which is the dependence unit used for inference.

\subsection{Validation Stability Across Training Seeds}

\begin{table}[h]
\centering
\caption{HybridContIn Validation stability under independently specified training seeds.}
\label{tab:app_seed_validation}
\begin{tabular}{rrrrrr}
\toprule
Seed & Params & Steps & Model NLL & LightGBM NLL & Model advantage \\
\midrule
17 & 0.9M & 5.5K & 3.1465 & 3.1756 & 0.0291 \\
29 & 0.9M & 5.5K & 3.1525 & 3.1756 & 0.0230 \\
43 & 0.9M & 5.5K & 3.1450 & 3.1756 & 0.0306 \\
\midrule
Mean & 0.9M & 5.5K & 3.1480 & 3.1756 & 0.0276 \\
Sample SD & 0.9M & 5.5K & 0.0040 & 0.0000 & 0.0040 \\
\bottomrule
\end{tabular}
\end{table}

\FloatBarrier
\subsection{Complete Multi-Seed Test Results}

\begin{table}[h]
\centering
\caption{All seed-period Test comparisons against the common identity-augmented LightGBM. Gain is \(\mathrm{NLL}_{\mathrm{LightGBM}}-\mathrm{NLL}_{\mathrm{VAIOM}}\); positive favors VAIOM.}
\label{tab:app_multiseed_test}
\resizebox{\textwidth}{!}{%
\begin{tabular}{lrrrrrrrr}
\toprule
Period & Seed & Params & Steps & Model NLL & LightGBM NLL & Mean gain & 95\% interval & Block win-rate \\
\midrule
2025H1 & 17 & 0.9M & 5.5K & 3.1764 & 3.2057 & +0.0292 & [+0.0205, +0.0406] & 84.8485\% \\
2025H1 & 29 & 0.9M & 5.5K & 3.1880 & 3.2057 & +0.0177 & [+0.0088, +0.0284] & 71.2121\% \\
2025H1 & 43 & 0.9M & 5.5K & 3.1817 & 3.2057 & +0.0239 & [+0.0157, +0.0341] & 81.8182\% \\
\midrule
2025H2 & 17 & 0.9M & 5.5K & 3.1359 & 3.1785 & +0.0426 & [+0.0291, +0.0626] & 92.4242\% \\
2025H2 & 29 & 0.9M & 5.5K & 3.1478 & 3.1785 & +0.0307 & [+0.0183, +0.0490] & 83.3333\% \\
2025H2 & 43 & 0.9M & 5.5K & 3.1396 & 3.1785 & +0.0388 & [+0.0273, +0.0558] & 92.4242\% \\
\bottomrule
\end{tabular}%
}
\end{table}

\FloatBarrier
The three-seed model mean is \(3.1820\pm0.0058\) in 2025H1 and \(3.1411\pm0.0061\) in 2025H2. All six paired intervals exclude zero.

\subsection{Canonical Tail-Bucket Decomposition}

\begin{table}[h]
\centering
\caption{Canonical seed-17 paired gain by tail group.}
\label{tab:app_tail_decomposition}
\begin{tabular}{lrrrrrr}
\toprule
Period & Params & Steps & Tail events & Negative tail & Positive tail & Combined tail \\
\midrule
2025H1 & 0.9M & 5.5K & 2,256 & +0.2545 & +0.1940 & +0.2251 \\
2025H2 & 0.9M & 5.5K & 2,279 & +0.3509 & +0.2428 & +0.3001 \\
2025 full year & 0.9M & 5.5K & 4,535 & +0.3036 & +0.2181 & +0.2628 \\
\bottomrule
\end{tabular}
\end{table}

\FloatBarrier
The entries are paired LightGBM-minus-VAIOM NLL differences in bits per event. They use realized buckets 1--4 for the negative tail and 13--16 for the positive tail.

\FloatBarrier
\FloatBarrier

\section{Statistical Inference Details}
\label{app:statistical_details}

\subsection{Asset-Month Block Bootstrap}

For each Test half, an observation block is one asset within one calendar month. The event-level paired loss difference is
\[
d_i=-\log_2 p_{\mathrm{LightGBM}}(y_i)+\log_2 p_{\mathrm{VAIOM}}(y_i),
\]
so positive values favor VAIOM. A bootstrap replicate samples the observed asset-month blocks with replacement until it contains the original number of blocks, concatenates all events in the sampled blocks and computes the event-weighted mean of \(d_i\). The reported 95\% interval is the 2.5th--97.5th percentile interval over 10,000 replicates generated with bootstrap seed 17.

The block win-rate is a separate descriptive statistic: the unweighted fraction of observed asset-month blocks whose within-block mean paired difference is positive. Its interval is obtained by applying the same block-resampling scheme to that block-level indicator.

\begin{table}[h]
\centering
\caption{Canonical fixed-model block-bootstrap inference.}
\label{tab:app_bootstrap_details}
\resizebox{\textwidth}{!}{%
\begin{tabular}{lrrrrrrrr}
\toprule
Period & Events & Blocks & Params & Steps & Replicates & Mean gain interval & Win-rate & Win-rate interval \\
\midrule
2025H1 & 33,068 & 66 & 0.9M & 5.5K & 10,000 & [+0.0205, +0.0406] & 84.8485\% & [75.7576\%, 92.4242\%] \\
2025H2 & 33,816 & 66 & 0.9M & 5.5K & 10,000 & [+0.0291, +0.0626] & 92.4242\% & [84.8485\%, 98.4848\%] \\
2025 full year & 66,884 & 132 & 0.9M & 5.5K & 10,000 & [+0.0273, +0.0476] & 88.6364\% & [82.5758\%, 93.9394\%] \\
\bottomrule
\end{tabular}%
}
\end{table}

\FloatBarrier
This procedure estimates Test-sample uncertainty conditional on a trained checkpoint and fixed baseline. Training uncertainty is addressed separately by the three independently trained seeds in Appendix~\ref{app:period_results}. No alternative block definition, Brier score, reliability diagram or calibration analysis is reported because those outputs are not present in the retained experiment record.

\FloatBarrier
\FloatBarrier

\section{Implementation and Reproducibility Details}
\label{app:implementation_details}

\subsection{Architecture Rungs and Tensor Shapes}

\begin{table}[h]
\centering
\caption{Predefined complete VAIOM architecture rungs.}
\label{tab:app_architecture_rungs}
\resizebox{\textwidth}{!}{%
\begin{tabular}{lrrrrrrr}
\toprule
Rung & Layers & Heads & Width & FFN width & Context & Total params & Non-positional $N$ \\
\midrule
0.9M & 4 & 4 & 128 & 512 & 512 & 890,086 & 824,550 \\
5M & 6 & 4 & 256 & 1,024 & 512 & 4,965,222 & 4,834,150 \\
15M & 8 & 6 & 384 & 1,536 & 512 & 14,726,850 & 14,530,242 \\
\bottomrule
\end{tabular}
}
\end{table}

\FloatBarrier
For pure ContIn, the input multilayer perceptron (MLP) has shape \(25\rightarrow128\rightarrow128\) with Gaussian Error Linear Unit (GELU) activation. HybridContIn concatenates this 128-dimensional projection with three 8-dimensional metadata embeddings, then applies a \(152\rightarrow128\rightarrow128\) fusion MLP. Learned absolute position embeddings are added afterward. Every Transformer block is pre-normalized and contains four-head causal self-attention and a \(128\rightarrow512\rightarrow128\) GELU feed-forward network in the selected rung.

The independent return head maps \(128\rightarrow16\). The selected MoMS head maps each hidden vector to four mixture logits and four separate 16-bucket component distributions; their probability-weighted sum is the final categorical return distribution. Gap and VolReg are categorical heads, while Ordinal uses 15 cumulative binary logits for the 16 ordered return buckets.

\subsection{Optimization and Numerical Evaluation}

\begin{table}[h]
\centering
\caption{Training and evaluation constants for the selected main configuration.}
\label{tab:app_optimization}
\begin{tabular}{p{0.43\textwidth}p{0.49\textwidth}}
\toprule
Item & Value \\
\midrule
Optimizer & AdamW \\
Learning rate / schedule & 0.0003 / constant; no warm-up \\
Adam coefficients / epsilon & \((0.9000,0.9990)\) / \(10^{-8}\) \\
Weight decay / dropout & 0.0100 / 0.1000 \\
Gradient clipping & Global norm 1.0000 \\
Micro-batch / accumulation / effective batch & 32 / 4 / 128 windows \\
Precision / parallelism & FP16 automatic mixed precision / data parallel \\
Checkpoint interval / selected step & 500 / 5,500 \\
Probability floor & \(10^{-12}\), followed by renormalization \\
\bottomrule
\end{tabular}
\end{table}

\FloatBarrier
\subsection{FullSeq Valid-Position Logic}

For a sampled asset window beginning at \(u\), the serialized base indicator is
\[
y_{a,s+1}>0
\land \neg\mathrm{mask\_any}_{a,s}
\land \neg\mathrm{target\_bad}_{a,s}.
\]
The implementation stores this conjunction as \texttt{valid\_base}. Return and Ordinal use it directly; Gap and VolReg intersect it with their own positive target-index masks. Each head then takes one global mean across all of its valid \((\text{window},\text{position})\) pairs in the current micro-batch. It does not first average each window. If no valid Gap or VolReg target exists in a micro-batch, that auxiliary term is skipped; return and Ordinal remain nonempty because every sampled window has a return-valid endpoint. Causal attention remains unchanged: the prediction at \(s\) can use only positions \(u{:}s\). Empty, padded, masked and invalid positions contribute to neither numerator nor denominator.

\subsection{LightGBM Hyperparameters}
\label{app:baseline_details}

LightGBM uses 1,000 estimators, 31 leaves, learning rate 0.0300 and \texttt{min\_child\_samples}=500. Regularization uses \(L_1=0.1000\), \(L_2=5.0000\), subsample 0.8000 and column subsampling 0.8000; the seed is 17. Early stopping uses patience 100 on a 20\% Train holdout with multiclass log loss. No hyperparameter search, class weighting or post-hoc calibration is applied. The fit is pooled across assets and capped at one million Train samples. A holdout gate rejects a fit whose NLL exceeds the Frequency baseline by more than 0.0200 bits per event. Predicted probabilities are clipped at \(10^{-12}\) and renormalized before evaluation.

For the main comparison, LightGBM receives the 25 numerical current-bar fields and three categorical identifiers. It receives no 512-bar sequence; the comparison therefore measures the combined value of sequence context and representation learning.

\clearpage
\bibliographystyle{plainnat}
\bibliography{references}

\end{document}